\documentclass[11pt, twocolumn]{article}
\usepackage[noblocks]{authblk} 
\usepackage[utf8]{inputenc}
\usepackage[english]{babel}
\usepackage{csquotes} 

\usepackage[style=numeric, sorting=none]{biblatex}
\usepackage[utf8]{inputenc}
\usepackage{xcolor, soul} 
\usepackage{amssymb}
\usepackage{amsmath}
\usepackage{amsthm} 
\usepackage{bm}
\usepackage{bbm} 
\usepackage{makecell} 
\usepackage{rotating} 
\usepackage{tablefootnote} 

\usepackage{placeins}
\usepackage{graphicx,caption} 
\usepackage{booktabs} 

\usepackage{multirow} 
\usepackage{changepage} 
\usepackage{subcaption}
\usepackage{mwe}

\usepackage{float}
\newfloat{example}{htbp}{loc}
\floatname{example}{Example}

\newfloat{rulepath}{htbp}{loc} 
\floatname{rulepath}{Rule}

\newcommand{\TTT}{\mathcal{T}}

\newcommand{\CCC}{\mathcal{C}}
\newcommand{\RRR}{\mathcal{R}}
\newcommand{\DDD}{\mathcal{D}}
\newcommand{\OOO}{\mathcal{O}}

\begin{document}

\date{} 

\author[1,2,*]{Klest Dedja}
\author[1,2]{Felipe Kenji Nakano}
\author[3]{Konstantinos Pliakos}
\author[1,2]{Celine Vens}
\affil[1]{Department of Primary Care, Kortrijk, KU Leuven}
\affil[2]{Itec, imec research group, Kortrijk, KU Leuven}
\affil[3]{Department of Management, Strategy, and Innovation, KU Leuven}
\affil[*]{corresponding author, emails are: \{firstname.lastname\}@kuleuven.be}

\title{BELLATREX: Building Explanations through a LocaLly AccuraTe Rule EXtractor}

\twocolumn[

 \begin{@twocolumnfalse}
    \maketitle
    \begin{abstract}
Tree-ensemble algorithms, such as random forest, are effective machine learning methods popular for their flexibility, high performance, and robustness to overfitting. However, since multiple learners are combined, they are not as interpretable as a single decision tree. In this work we propose a novel method that is Building Explanations through a LocalLy AccuraTe Rule EXtractor (Bellatrex), and is able to explain the forest prediction for a given test instance with only a few diverse rules.
Starting from the decision trees generated by a random forest, our method 1) pre-selects a subset of the rules used to make the prediction, 2) creates a vector representation of such rules, 3) projects them to a low-dimensional space, 4) clusters such representations to pick a rule from each cluster to explain the instance prediction.
We test the effectiveness of Bellatrex on 89 real-world datasets and we demonstrate the validity of our method for binary classification, regression, multi-label classification and time-to-event tasks. To the best of our knowledge, it is the first time that an interpretability toolbox can handle all these tasks within the same framework. We also show that our extracted surrogate model can approximate the performance of the corresponding ensemble model in all considered tasks, while selecting only few trees from the whole forest. We also show that our proposed approach substantially outperforms other explainable methods in terms of predictive performance.
    \end{abstract}
    

	\vspace{0.5cm}
 \end{@twocolumnfalse}
 ]


\theoremstyle{definition}  
\newtheorem{definition}{Definition}[section]
\newtheorem*{definition*}{Definition}
\newtheorem*{remark}{Remark}

\section{Introduction} \label{sec:Intro}

Machine Learning (ML) models are nowadays employed in various domains with excellent performance. When the performance is key to having a competitive advantage, ML models are widespread even when the algorithmic procedure is too complicated to be understood by humans (so called ``black box" algorithms). However, in fields where the stakes for a single decision are high (healthcare, banking, etc.)~\cite{Cutillo2020} there is still scepticism in adopting such models.
Professionals are accountable for any costly mistake, and in such cases they want to fully understand and trust the outcomes of a ML model. An important element that leads to trust in ML models is to achieve more explainability in ML~\cite{Dzindolet2003}.

In recent years, the ML community has therefore been developing tools that help ML practitioners and end-users to understand the predictions of black box models, leading to a booming of the field of Explainable AI (X-AI). 

As a result, X-AI methods nowadays make up a vast literature and can be categorised in several ways. One way is to distinguish the type of output explanations: \emph{global} explanations~\cite{Friedman2001, Goldstein2015}, where the aim is to identify trends in the data and explain the model as a whole, or \emph{local} explanations \cite{Ribeiro2016, Lundberg2017}, where the aim is to understand why a certain prediction is made for a given instance.
Another common differentiation is made based on whether 
the method has access to the internal components of the model and its architecture.
If that is the case, we talk about a \emph{model-specific} approach, as opposed to a \emph{model-agnostic}~(e.g.~\cite{Ribeiro2016, Lundberg2017}) approach.


In this study, we propose Bellatrex: a local, model-specific method that condenses the predictions made by a random forest (RF)~\cite{Breiman2001} into a few rules. By doing so, our method uses information directly extracted from the original RF. Furthermore, the procedure is optimised to follow closely the RF predictions for every instance, and maintains the original RF high performance.
In addition to high performance, Bellatrex provides different insights by selecting rules that are dissimilar from each other. Furthermore, given its local method, the extracted rules are different for each instance.

The choice to specifically explain RF predictions is driven by the fact that RF has a great predictive performance and is 
easier to train than other highly performing methods, such as deep neural networks. Moreover, RF has been extended to a wide range of learning tasks, including binary classification, multi-label classification, regression and time-to-event prediction. We should highlight that our method is designed in such a way that it is independent of the learning task at hand, and consequently, it is capable of handling all these scenarios.

Bellatrex is built on top of a trained RF and consists of four main steps. First, for a given input instance, a subset of rules is selected based on how close their prediction is with respect to the original ensemble. Second, a vector representation is built by extracting information from the rule structure. Third, dimensionality reduction of such representations is performed and lastly, we apply clustering on the projected vector representations, extracting this way only a few prototype rules. These final rules are then presented to the end user as an explanation for the forest prediction. By keeping the number of prototype rules low, we provide a few diverse explanations, which can be inspected by the user.

Preliminary results of this work were presented in an earlier workshop paper~\cite{Dedja2021} and were restricted to binary and time-to-event predictions. Here, we extend this work to regression (both single- and multi-target) and multi-label classification. Furthermore, we also provide a thorough experimental evaluation of our method against competitor methods from the literature in predictive performance, explanation complexity and explanation diversity, followed by an ablation study which validates the steps in our method. Overall, our contributions to the field of X-AI can be summarised as follows:
\begin{itemize}
    \item We propose a local approach for explaining RF predictions, with a method that extracts a few meaningful decision paths from a trained RF. 
    Importantly, the extracted rules are tailored to the instance to be explained, which represents, to the best of our knowledge, a novelty in the field of explaining RF predictions.
    \item We evaluate our local explainer in five different scenarios: binary classification, single- and multi-target regression, multi-label classification and time-to-event prediction, where the latter three are seldom explored in existing explainability methods. The inclusion of the time-to-event analysis in particular, highlights a novelty of our work since, to the best of our knowledge, no existing model-specific interpretability method is compatible with an adaptation of RF to time-to-event scenarios.
    \item We make our method available to the ML community by providing a Python implementation through GitHub\footnote{https://github.com/Klest94/Bellatrex}, compatible with the scikit-learn and scikit-survival Python libraries. 
\end{itemize}

The remainder of the manuscript is organised as follows: Section~\ref{sec:Related-work} positions our work within the framework of post-hoc, RF-specific explainers, and underlines the differences with the existing literature; Section~\ref{sec:Background} provides the background information for the method section (Section~\ref{sec:Method}), where our model is introduced. These are followed by Section~\ref{sec:Experiments}, dedicated to the experimental set-up and Section~\ref{sec:Results}, where the predictive performance of the method, the average complexity of the extracted rules, and the average dissimilarity of the latter are reported together with the outcomes of the ablation study. Next, we provide two examples of explanations generated by Bellatrex in Section~\ref{sec:Examples}. Finally, we draw our conclusions and share future directions of this work (Section~\ref{sec:Conclusion}).

\section{Related work} \label{sec:Related-work}

As previously mentioned, Bellatrex falls in the category of local, model-specific methods, and in particular, Bellatrex explains the outcomes of a trained random (survival) forest model in a post-hoc fashion.
Such family of RF-specific methods has produced a rich literature as shown in~\cite{Aria2021, Haddouchi2019}, and can be categorised 
in three sub-categories, namely: \emph{size reduction}, \emph{rule extraction}, and \emph{local explanation}. 
In this section, we discuss other model-specific approaches that use RFs as the underlying black-box model.

The earliest example of such is~\cite{Chipman1998}. This work proposed a procedure for analysing a RF in binary classification tasks, and concepts such as \emph{tree-metric}, a distance metric defined on the space of tree learners, are also introduced.
Trees are then represented in a two-dimensional space by using Multi Dimensional Scaling~\cite{Mead1992} on the tree-distance matrix. The authors then perform a qualitative study and mention the possibility of clustering the resulting tree representations.

A more recent work by~\cite{Sies2020} followed the path traced by~\cite{Chipman1998} by suggesting a range of tree similarity metrics and by performing clustering on such tree representations. Furthermore, the authors implicitly propose a vector representation of trees based on the covariates used to build them and it consists of an adaptation of the approach from~\cite{Banerjee2012}. 
Given such representation, the authors clustered and extracted a few trees out of a random forest and illustrate the validity of their framework ``C443" on simulated and real world datasets.
The resulting method has some features in common with ours, namely, the idea of representing tree- (or rule-) distances for clustering purposes. However, the focus in C443 lays more on giving an overview of the possible approaches rather than optimising the performance of a surrogate model. Furthermore, the C443 method aims at global explanations (e.g. by looking at trends and sources of heterogeneity), rather than focusing at an instance-base level.


Later work by~\cite{Zhao2019} (``iForest") made use of a tree-distance metric intuition to project tree paths via t-SNE~\cite{hinton2002}, and made also use of visual analytics to improve interpretability. These visuals include a ``feature view" that shows the relationship between a selected covariate and a prediction (similarly to a Partial Dependency plot~\cite{Friedman2001}) as well as a control panel that just lets the user see the decision paths that lead to the prediction of a group of instances. Unlike Bellatrex, there is no rule extraction or forest approximation mechanism and none of the original learners are shown to the end user. 


Other relevant approaches include~\cite{Moore2018}, where the authors presented a method for computing how a covariate within a value range influences the final prediction of a model, while an optimisation-based approach that exploits the bijective relationship between a tree leaf and a subspace of the input space is proposed by~\cite{Hara2018}.
Similarly, recent work~\cite{LionForest, multilabel-LionForest} proposes a method called ``LionForest", where rules are generated in a local approach fashion, and path reduction approaches are run in order to provide shorter rules with more ``conclusive" (i.e., stable) explanations. 

A toolbox named ``SIRUS" is proposed in~\cite{Benard2021}, where a set of rules is extracted from an RF and shown as an explanation. 
More specifically, rules are first generated by large RF and a pre-selection step picks the most commonly occurring ones. After that, a ``post-treatment" step narrows down the selection by eliminating rules associated to a linear combination of other rules of higher frequency. The procedure stops when the desired number of final rules is reached and is then shown to the end user. These final rules serve as an explanation of the original RF and their average prediction can be used as a the prediction of the obtained surrogate model.
The resulting explanation is global and consists of a limited number of rules that are stable under different subsets of data, and 
whose outcome can be averaged to get a prediction.

More recently, ``RuleCOSI+"~\cite{RuleCOSI+} has been proposed as another global, rule-extraction (according to the categorisation by~\cite{Aria2021}) method that greedily combines and simplifies the corresponding base trees.
More specifically, the algorithm builds a decision list by iteratively creating new rules by merging or pruning the ones generated by the tree ensemble. At the end of each iteration step, the generated rule is added to the existing list under the condition that it is beneficial to the generalisation performance.
When no extra rules can be added, an empty rule (playing the part of an ``else" instance) is appended, and the final model consists of the set of rules found, structured as members of a decision list. That is, for a given instance to be explained, the rules are evaluated in order of appearance until one of them is found to hold and is used as the models' prediction. If no rules are fired, the final (``empty") rule is used to make the prediction. 

Finally, a size-reduction, global approach is represented by ``Hierarchical Shrinkage"~\cite{HS2022} (HS) algorithm. The idea is to increase performance and robustness of tree learners by shrinking the predictions of any leaf node towards the predictions of its ancestors. The shrinking operation is performed repeatedly and is controlled by a regularisation parameter. The resulting tree has a size that can be controlled and offers a global explanation to the model. 

It is worth mentioning that the aforementioned RF-specific, post-hoc explainability methods follow a global approach and therefore show the same explanation regardless of the input instance. Bellatrex, on the other side, follows a local approach that adapts its explanations to the instance of interest. This leads to an increased flexibility of the output explanation and allows Bellatrex to closely follow the predictions of the underlying RF.

\section{Background} \label{sec:Background}

Random forest~\cite{Breiman2001} (RF) is a computationally efficient ML method that delivers excellent predictive performance, and since its appearance, RFs have been extremely successful in performing classification and regression tasks. Moreover, RFs have been extended to a variety of other learning tasks, including multi-label classification~\cite{Vens2008} and survival analysis~\cite{Ishwaran2008}.
Most of the extensions that have been proposed over the years are associated with the node splitting mechanism as well as the prototype function (i.e., the function that provides predictions in the leaves), while less modifications have been proposed for other RF characteristics, such as the bootstrapping.

A common splitting criterion for RFs in binary classification is the Gini impurity index, whereas a common prototype function returns the most frequent class label among training instances falling in the leaf. For single-output regression tasks, the splitting rule that maximally reduces within-node variance in the target values is usually chosen, and the average target value is returned as prediction.

In multi-label classification, a single instance may belong to multiple classes and, hence, the task is to predict the correct subset of class labels for each instance. Multi-label classification has many applications, including text classification, image annotation, protein function prediction, etc. Several machine learning methods have been adapted to multi-label classification, including random forests~\cite{Kocev2013, Pedregosa2011}. In order to take into account the multiple labels, the splitting criterion is adapted by maximising either the average impurity reduction~\cite{Pedregosa2011} or average variance reduction~\cite{Vens2008}, across all the (binary) labels. These criteria implicitly take into account label dependence and are often preferred to methods that train output labels independently~\cite{Kocev2013}. In both cases, the leaves return a vector, where each component represents the proportion of instances annotated with the corresponding label.

Similarly, RF have been adapted to multi-target regression tasks, where multiple continuous targets are predicted at once. In this case, the reduction in within-node variance is computed for all targets and the average is taken.


Survival analysis is a branch of statistics that has only recently been explored in the machine learning community~\cite{Wang2019}. The goal of such analysis is to predict the time until an event occurs (hence, this task is also known as time-to-event prediction). Applications include clinical studies, where the outcome of interest is the time until a patient dies, suffers from complications, is discharged from hospital, etc.  A key challenge in this field is the so called \emph{censoring} effect, that is when the exact time of the event is not observed, resulting in partial information. For instance, in the case of right censoring, which is the type of censoring included in our study, the recorded event time is an underestimate of the true time. This may for instance occur when a patient drops out from the study or has not reached the outcome event before the study ends. Decision trees and random forests (as well as other machine learning methods) have been adapted to survival data, they are called (Decision) Survival Trees~\cite{LeBlanc1993} 
and Random Survival Forests (RSF)~\cite{Ishwaran2008}, respectively. The main splitting criterion being used is the \emph{logrank} score~\cite{Peto1972}, a criterion that guarantees good generalisation for censored time-to-event data~\cite{Ishwaran2008}.
This criterion aims at splitting the population in maximally different subgroups with maximally different hazard functions. 
At each leaf node, a Kaplan-Meyer~\cite{KaplanMeyer1958} estimate of the survival function for the population that falls in that node is built.  

\section{Proposed Method} \label{sec:Method}

\begin{figure*}[ht]
    \centering
    \includegraphics[width=0.9\textwidth]{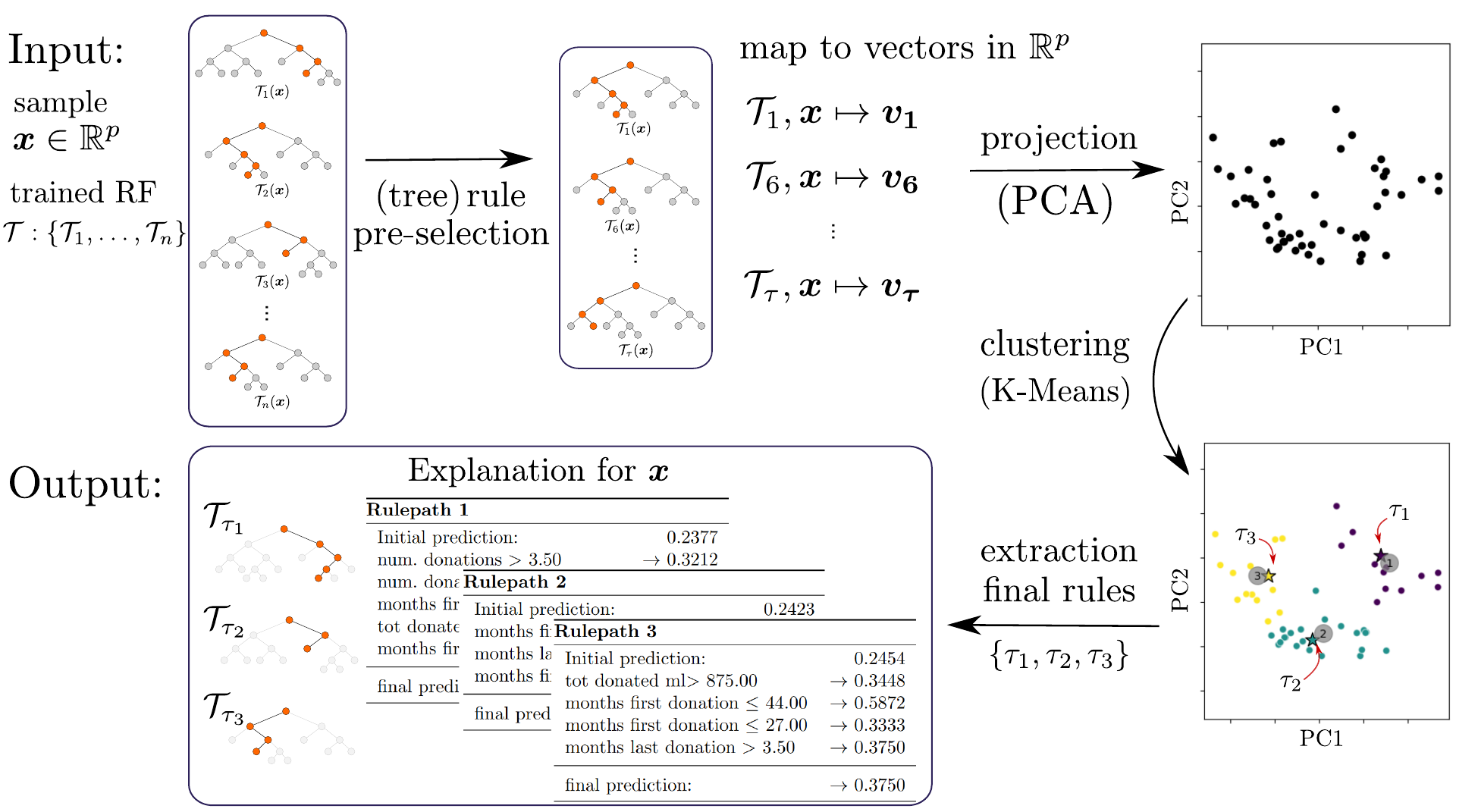}
    \caption{Schematic representation of the steps in Bellatrex. Given a trained RF and an instance (top left),  a pre-selection is performed (top), followed by the vectorisation step. Such vectors are projected (top right) and clustered (bottom right), the final selected rules are shown to the end-user (bottom left).}
    \label{fig:method}
\end{figure*}

Let $\TTT$ be a random forest and let $\bm{x}$ be an instance for which we want to find an explanation. Bellatrex uses these inputs to generate an explanation, and an overview of the procedure is illustrated in Figure~\ref{fig:method}. 

We first extract $\tau$ trees $\TTT_i \in \TTT$ that generate the most similar predictions $\TTT_i(\bm{x})$ compared to the ensemble model prediction $\hat{y} = \TTT(\bm{x})$. Next, we represent each of the selected trees $\TTT_i$ as a vector; this vector can be a function of the tree structure, or solely depend on the path traversed by $\bm{x}$. Only a few studies~\cite{Chipman1998, Sies2020} have addressed the issue of representing decision trees as a numerical vector, and they only focus on global representations.
Given the local nature of Bellatrex, we propose two path-based approaches, where we follow the instance of interest $\bm{x}$ as it traverses the tree and record the input covariates used to perform each split. 
In the first approach, which we call \emph{simple}, the vector representation $[f_1, f_2, \dots, f_p]$ consists of the number of times each of the $p$ covariates is selected to perform a split along the path traversed by~$\bm{x}$. In formulas:
\begin{equation} \label{eq:path-based}
\begin{split}
\TTT_i, \bm{x} \mapsto &  \bm{f} =[ f_{1}, f_{2}, \dots, f_{p}], \text{ where} \\
f_i = & \sum_{k \in \text{path$(\bm{x})$}} \mathbbm{1} \{\text{split on covariate $i$}\}.
\end{split}
\end{equation}
Unlike the tree-based approach that arises from~\cite{Sies2020}, the path-based vector representation of $\TTT_i$ depends on the instance of interest $\bm{x}$ and is therefore more tailored to the local nature of Bellatrex. 

A drawback of this representation is that splits have the same contribution regardless of the importance they have in the final predictions, whereas splits performed close to the root node should be considered as more important than the ones close to a leaf. To tackle this, we also propose a \emph{weighted} version of the previous approach, where the contribution of each node-split $k$ is weighted by the proportion of instances $\omega_k$ traversing the node.
In formulas, the representation becomes:
\begin{equation} \label{eq:weighted-path-based}
\begin{split}
\TTT_i, \bm{x} \mapsto & \bm{g} =[g_{1}, g_{2}, \dots, g_p], \text{ where} \\
 g_i = & \sum_{k \in \text{path$(\bm{x})$}} \omega_k \cdot \mathbbm{1}\{ \text{split on covariate $i$}\}.
\end{split}
\end{equation}
It is worth noting that the resulting weights are positive and upper bounded by 1, which is the weight assigned to the root node split, furthermore, the assigned weights decrease as the depth increases. By doing so, rules that differ (only) in the root node splitting covariate will be further apart in Euclidean distance compared to the ones whose decision paths differ (only) in the deeper nodes, which is in agreement with common intuition.



Next, we project such vector representations to a low-dimensional space using Principal Component Analysis (PCA)~\cite{Pearson1901}. The idea is to remove the noise, to improve computational efficiency for later steps, and to enable a better visualisation of the subsequent clustering. The representations are real-valued vectors of length $d$ at this stage, where $d$ is the number of output dimensions of the PCA.

\begin{figure}[ht]
    \centering
    \includegraphics[width=0.95\columnwidth]{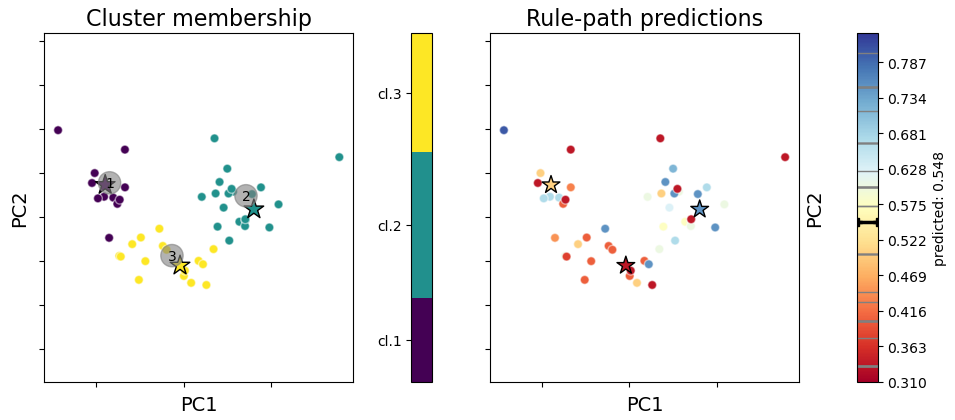}
    \caption{Left side: Example of the clustering step following rule pre-selection, vectorisation and PCA for the ``blood" dataset (Section~\ref{sec:performances} for details) and a test instance $\bm{x}$. The centroids of the K-Means++ algorithm are shown in grey, and the final representative rules are plotted with a star shape. Right side: individual rule predictions and Bellatrex final prediction for $\bm{x}$.}
    \label{fig:plot-clusters}
\end{figure}

In order to obtain a diverse set of explanations, as the next step, we perform clustering on the vector representations using a standard clustering method, such as K-Means++~\cite{vassilvitskii2006} (see an example in Figure~\ref{fig:plot-clusters}).
By doing so, we group the vectors into $K$ clusters, we identify the vector closest to each cluster centre and pick the corresponding rule as a representative for explaining the outcome of the model. The rules extracted with this procedure correspond to what we call \emph{final rules}, and the nature of the partitioning step guarantees the distance among the vector representations 
of the selected final rules is maximal.
It is worth noting that similar results can be obtained with other partitioning algorithms such as K-median~\cite{park2009}, as long as the aim of the algorithm is to maximise between-cluster distance. 

Finally, given the $K$ clusters, the corresponding final rules $\TTT_{\tau_k}$, and the instance $\bm{x}$, we build a surrogate model prediction $\bar{y}$ as follows:
\begin{equation} \label{eq:weighted-avg}
    \bar{y} = \sum_{k=1}^K w_k \TTT_{\tau_k}(\bm{x})
\end{equation}
where $w_k$ represents the weight given to the cluster $k$. We define $w_k$ as the proportion of the $\tau$ rules (selected in the first step) that are part of the cluster. It follows that $\sum w_k = 1$, and that the surrogate model predicts a weighted average of the selected rules.

Our method requires three hyperparameters: the number of trees $\tau$ to keep in the pre-selection phase, the number of output components $d$ for PCA, and the number of clusters $K$. The optimal values can be tuned for each test instance separately, by calculating the \textit{fidelity} of the surrogate model to the original ensemble.
This measure is an indicator of how closely the original ensemble prediction is being imitated by our surrogate model, and has recently been listed~\cite{Guidotti2018} as a desirable property for interpreting black-box models. Given an instance $\bm{x}$, the prediction made by the original R(S)F $\hat{y}$ and the new surrogate prediction $\bar{y}$, we define and compute fidelity $\mathcal{F}(\bm{x})$ as:
\begin{equation} \label{eq:fidelity}
\mathcal{F}(\bm{x}) = 1 - \Vert \hat{y} - \bar{y} \Vert_2.
\end{equation}
Depending on the predictive scenario, $\hat{y}$ and $\bar{y}$ are either scalars for predicted probabilities (binary classification), estimated scalar values (regression), vectors of predicted class probabilities (multi-label classification) or real valued-vectors (multi-target regression), or scalar predicted risk scores (survival analysis).
Note that $\mathcal{F}(\bm{x}) = 1$ for a given instance $\bm{x}$ indicates perfect fidelity of Bellatrex to the original model. 
As a result, the hyper-parameter combination $(\bar{K}, \bar{d}, \bar{\tau})$ that maximises $\mathcal{F}$ can be chosen. 

An important advantage of our method is that it does not require an external validation set to tune these hyperparameters; the proposed method looks instead at the \emph{predicted} labels from the underlying RF model which at this stage serves as an oracle, in a procedure similar to the one described in~\cite{Zhou2016}. The hyperparameters that, for a given test instance, yield the prediction closest to the underlying RF prediction are the ones selected. By doing so, we increase the fidelity of the surrogate model to the underlying black box, and we concurrently allow a more efficient use of data, since a greater fraction of it can be used to train the underlying model instead of being left aside for the purpose of hyperparameter tuning.

Alternatively, the parameter values can be determined by the user. In the latter case, end-users have the option to choose, for example, the number of clusters and thus, explore the trade-off 
between having a simple (single) explanation against obtaining multiple (different) explanations for a given example.

\subsection{Computational complexity}

We now analyse the computational complexity of Bellatrex. 
Given the number of data instances $n$, the number of covariates $p$, and the number of learners $m$, the computational complexity of Bellatrex can be estimated by looking at the complexity of its four main steps:
\begin{itemize}
    \item The sorting algorithm for rule pre-selection, with complexity $\OOO(m\log m)$;
    \item The vectorisation process of the pre-selected rules: $\OOO(m \log n)$;
    \item The dimensionality reduction step, performed with PCA, with $\OOO(n_{max}^2 n_{min})$~\cite{PCA}, where $n_{max}=\max(m,p)$ and $n_{min}=\min(m,p)$.
    \item partitioning algorithm, namely K-Means++, with $\OOO(p m)$, followed by a nearest neighbour search, performed with KDTree~\cite{KDTree}, with $\OOO(p m \log(m))$.
\end{itemize}
We conclude that the computational complexity of Bellatrex is mainly driven by the dimensionality reduction step. 


Translating the above analysis into practice, we observe that querying Bellatrex for an explanation

 is quite fast and takes a couple of seconds to run on a laptop. 
 Moreover, we notice that Bellatrex is sensitive to dataset size, but not so much to the 
 number of trees $m$ of the underlying RF learner, which is typically $m=100$.

\section{Experiments} \label{sec:Experiments}

We evaluate our method (with simple and weighted vector representations) across multiple datasets; unless otherwise specified, the datasets are selected from publicly available repositories such as UCI\footnote{https://archive.ics.uci.edu/ml/datasets.php} and MULAN\footnote{http://mulan.sourceforge.net/datasets-mlc.html} or downloaded from \textsf{sklearn}'s library.
Overall, we include as many as 89 datasets spanning five different prediction scenarios, more specifically:
\begin{itemize} \setlength{\itemsep}{0pt}
    \item 24 datasets for binary classification;
    \item 14 datasets for survival analysis (time-to-event prediction);
    \item 19 datasets for regression;
    \item 13 datasets for multi-label classification;
    \item 19 datasets for multi-target regression.
\end{itemize}

\subsection{Data and pre-processing}

We report an overview of the datasets used for our study, 
where information about the number of data instances, the number of covariates, and the format of the output labels is included. For the binary classification case, the frequency of the positive class is reported, whereas for the time-to-event data we report the censoring rate. 
For tasks involving a vector prediction (multi-label classification and multi-target regression), we report the number of labels (or targets) to be predicted.

A number of pre-processing steps were performed before running the experiments in Section~\ref{sec:Experiments-set-up}.
More specifically, categorical variables were one-hot encoded, instances and covariates with more than 30\%~missing values were dropped, and the remaining missing values were imputed with MICE~\cite{MICE, VanBuuren2007}. The resulting cleaned datasets properties are shown in Tables~\ref{tab:binary-info}-\ref{tab:mtr-info}.

Table~\ref{tab:binary-info} shows the binary classification scenario, whereas Table~\ref{tab:multi-label-info} shows its multi-label counterpart, where each instance can be associated to multiple labels at the same time.
\begin{table}[!ht]
    \scriptsize
    \begin{center}
    \captionsetup{width=.9\linewidth}
    \caption{Overview of the datasets for binary classification tasks}\label{tab:binary-info}
    \begin{tabular}{l|cc}
    \toprule
    & \makecell{data size \\ $(n , p)$} & \makecell{\% positive \\ class} \\
    \midrule
    blood &(748 , 4) &24 \\
    breast cancer diagn. &(569, 30) &37 \\
    breast cancer &(699, 10) &34 \\
    breast cancer progn. &(198, 33) &24 \\
    breast cancer coimba &(116, 9) &55 \\
    colonoscopy Green &(98, 62) &68 \\
    colonoscopy Hinselm. &(97, 62) &85 \\
    colonoscopy Schiller &(92, 62) &73 \\
    divorce &(170, 54) &49 \\
    Flowmeters &(87, 36) &60 \\
    haberman &(306, 3) &26 \\
    hcc-survival &(165, 49) &62 \\
    ionosphere &(351, 34) &36 \\
    LSVT voice &(126, 310) &67 \\
    mamographic &(961, 5) &46 \\
    musk &(476, 166) &43 \\
    parkinson &(756, 753) &75 \\
    risk factors &(858, 35) &6  \\
    crashes &(540, 18) &91 \\
    sonar &(208, 60) &53 \\
    SPECT &(267, 22) &79 \\
    SPECTF &(267, 44) &79  \\
    vertebral &(310, 6) &68 \\
    wholesale &(440, 7) &32 \\
    \bottomrule
    \end{tabular}
    \footnotetext[1]{short for ``breast cancer"}
    \footnotetext[2]{short for ``colonoscopy"}
    \end{center}
\end{table}

\begin{table}[ht]
    \scriptsize
    \begin{center}
    \captionsetup{width=.9\linewidth}
    \caption{Overview of the datasets for multi label classification tasks}\label{tab:multi-label-info}
    \begin{tabular}{l|cc}
    \toprule
    & \makecell{data size \\ $(n , p)$} & \makecell{number \\ of labels} \\
    \midrule
    birds &(645, 260) &19 \\
    CAL500 &(502, 68) &174 \\
    emotions &(593, 72) &6 \\
    enron &(1702, 1001) &53 \\
    flags &(193, 19) &7 \\
    genbase &(662, 1185) &27 \\
    langlog &(1460, 1004) &75 \\
    medical &(978, 1449) &45 \\
    ng20 &(4379, 1006) &20 \\
    scene &(2407, 294) &6 \\
    slashdot &(3782, 1079) &22 \\
    stackex chess &(1675, 585) & 227 \\
    yeast &(2417, 103) &14 \\
    \bottomrule
    \end{tabular}
    \end{center}
\end{table}
The datasets used in the framework of the time-to-event scenario are shown in Table~\ref{tab:survival-info}.

\begin{table}[!ht]
    \scriptsize
\begin{center}
\caption{Overview of the time-to-event datasets.}\label{tab:survival-info}
\begin{tabular}{l|cc}
\toprule
 & \makecell{data size \\ $(n , p)$} & \makecell{censoring \\ rate (\%)}  \\
\midrule
addicts &(238, 3) &37 \\
breast cancer survival &(198, 80) &74 \\
DBCD &(295, 4919) &73 \\
DLBCL &(240, 7399) &43 \\
echocardiogram &(130, 9) &32 \\
FLChain &(7874, 8) &72 \\
gbsg2 &(686, 8) &56 \\
lung &(228, 8) &28 \\
NHANES I\tablefootnote{https://www.cdc.gov/nchs/nhanes/about\_nhanes.html}  &(9931, 18) &65 \\
PBC  &(403, 19) &56 \\
rotterdam (excl. \textsf{recurr}\tablefootnote{the \textsf{reccur} covariate is highly correlated with the final time to event}) &(2982, 11) &57 \\
rotterdam (incl. \textsf{recurr}) &(2982, 12) &57 \\
veteran &(137, 9) &7 \\
whas500 &(500, 14) &57 \\
\bottomrule
\end{tabular}
\footnotetext[2]{https://www.cdc.gov/nchs/nhanes/about\_nhanes.html}
\footnotetext[3]{the \textsf{reccur} covariate is highly correlated with the final time to event}
\end{center}
\end{table}

Finally, we consider the single target and multi-target regression scenarios. For a better comparison of prediction errors, all target covariates have been normalised to the $[0, 1]$ interval. The employed datasets are shown in Tables~\ref{tab:regression-info}-\ref{tab:mtr-info}.

\begin{table}[ht]
    \scriptsize
    \centering
    \caption{Overview of datasets with regression tasks.}\label{tab:regression-info}
    \begin{tabular}{l|cc}
    \toprule
    & \makecell{data size \\ $(n , p)$} \\
    \midrule
    airfoil &(1503, 5) \\
    Ames Housing &(2915, 283) \\
    auto mpg &(398, 7) \\
    bike sharing &(731, 12) \\
    boston housing &(506, 14) \\
    california &(20640, 8) \\
    car imports &(201, 63) \\
    Computer&(209, 6) \\
    concrete compress &(1030, 8) \\
    concrete slump &(103, 9) \\
    ENB2012 cooling &(768, 8) \\
    ENB2012 heating &(768, 8) \\
    forest fires &(517, 12) \\
    PRSA data &(41757, 13) \\
    slump dataset &(103, 9) \\
    students maths &(395, 43) \\
    wine quality all &(6497, 12) \\
    wine quality red &(1599, 11) \\
    wine quality white &(4898, 11) \\
    \bottomrule
    \end{tabular}
\end{table}

\begin{table}[ht]
    \scriptsize
    \caption{Overview of datasets with multi-target regression tasks.}\label{tab:mtr-info}
    \begin{tabular}{l|cc}
    \toprule
    & \makecell{data size \\ $(n , p)$} & \makecell{number \\ of targets} \\
    \midrule
    andro &(49, 30) &6 \\
    atp1d &(337, 370) &6 \\
    atp7d &(296, 370) &6 \\
    edm &(154, 16) &2 \\
    enb &(768, 8) &2 \\
    ENB2012 &(768, 8) &2 \\
    jura &(359, 15) &3 \\
    oes10 &(403, 298) &16 \\
    oes97 &(334, 263) &16 \\
    osales &(639, 401) &12 \\
    rf1 &(9125, 64) &8 \\
    rf2 &(4332, 576) &8 \\
    scm1d &(9803, 280) &16 \\
    scm20d &(8966, 61) &16 \\
    scpf &(1137, 23) &3 \\
    sf1 &(323, 10) &3 \\
    sf2 &(1066, 9) &3 \\
    slump &(103, 7) &3 \\
    wq &(1060, 16) &14 \\
    \bottomrule
    \end{tabular}
\end{table}

\subsection{Competing methods} \label{sec:Competing-methods}

The proposed algorithm is compared against several competing methods. More specifically we include:
\begin{itemize}
    \item Four tree-based models, of various degree of performance and interpretability.
    \item Two linear models that are ante-hoc interpretable
    \item Four methods from literature that provide post-hoc explanations to RF predictions
\end{itemize}

\subsubsection*{Tree-based methods} \label{sec:tree-based}

The first family of models that we compare against includes tree-based learners 
such as DT and RF. These methods represent the two extremes of the performance-interpretability trade-off.
In addition, we propose two approaches, namely ``Small R(S)F" and ``OOB (Survival) Trees", that position themselves somewhere in between.

The Small R(S)F method consists of training a Random (Survival) Forest with $K$ base tree learners, where the value of $K$ is chosen for each test instance to be equal to the one used by Bellatrex\footnote{For reporting predictive performance, in the interest of space, $K$ is set as in the weighted path-based Bellatrex, and similar values are observed if the unweighted path-based set-up was used instead.}. 


The OOB (Survival) Trees method consists of averaging the prediction of $K$ selected trees from a full R(S)F, where the $K$ trees with smallest out-of-bag (OOB) error are selected. The value $K$ is again set to the same value as the number of final rules extracted by Bellatrex (in its weighted set-up).




\subsubsection*{Linear, ante-hoc explainable methods}

Next, we perform a performance comparison with competing methods that are \emph{ante-hoc} explainable, that is, models that are interpretable by design. We include a regularised regression model (LR) with elastic net penalty for binary classification, multi-label classification, regression, and multi-target regression. Additionally, we consider a regularised Cox-Proportional Hazard (Cox-PH)~\cite{cox1972} for time-to-event data.

\subsubsection*{Post-hoc competitors}

Finally, we compare our method against recent work in literature. For this purpose we select from the related work Section~\ref{sec:Related-work} the methods that have a publicly available source code. The selection narrows down to C443~\cite{Sies2020}, SIRUS~\cite{Benard2021}, RuleCOSI+~\cite{RuleCOSI+} and HS~\cite{HS2022}. It is worth mentioning that only SIRUS and HS fully support regression tasks, and that only the latter is adapted to multi-target regression and multi-label classification. Moreover, none of the selected competing methods is adapted to a survival analysis scenario.

To begin with, to set up a comparison between C443 (available for binary classification tasks only, within the homonym package in \textsf{R}) and Bellatrex, we train the underlying RF black-box model with the same number of learners and stopping criterion as in Section~\ref{sec:Experiments-set-up}. Next, we run C443, collect the output trees, and average their prediction (weighted proportionally to cluster size) for a close comparison. As for the number of clusters 
we run C443 with $K=3$, which is the upper bound for the number of rules extracted by Bellatrex.

To run the comparison with SIRUS, we keep the parameters suggested by the authors. More specifically, the underlying RF model is trained with 5000 trees and the number of final rules to be extracted is set to 10.

For RuleCOSI+, we run the proposed method 
after increasing the maximum rule length to 5 and the rule confidence to be at least 0.9, we do so to achieve better performance. We therefore report RuleCOSI+ results for this particular choice of hyperparameters.

Finally, we consider Hierarchical Shrinkage~\cite{HS2022} (HS) method in its default configuration. Such configuration sets the maximum number of leaves of the underlying DT learner to 20.



\subsection{Evaluation metrics}

We consider several metrics to validate our method's performance and interpretability, as well as to compare it against similar methods in literature.
Firstly, we look at predictive performance, whoch is also a desirable property for model interpretability~\cite{Guidotti2018, Doshi-Velez-2017}. Next, we look at the complexity of the explanations generated by our method. Finally,  we check against redundancy of the explanations by looking at the dissimilarity of the generated rulesets.

\subsubsection*{Performance}

The predictive performance of the models is computed by means of commonly used measures in each of the five tasks. More specifically, we evaluate the 
AUROC for binary classification, the weighted average AUROC in multi-label classification (averaged over each label with weight proportional to the number of positive instances of the label), 
mean absolute error
(MAE) for single target and multi-target regression tasks (in the latter case averaged over the targets), and concordance index (C-index) for the survival analysis data.

\subsubsection*{Complexity} \label{sec:complexity}

Next to predictive performance, we investigate the complexity of the generated explanations, a concept that is inversely related to model interpretability. The notion of interpretability is itself difficult to measure objectively as no general definition of interpretability exists~\cite{Molnar2020, Doshi-Velez-2017}, and many metrics have been proposed~\cite{Hoffman2018}.
However, when it comes to specifically interpreting tree-based models, 
it is common to report the total number of rule-splits shown by the model explainer as a measure of interpretability~\cite{RuleCOSI+, Hara2018, inTrees}, where a lower number of rule-splits (complexity) corresponds to a higher interpretability.
Formally, consider an explanation made of rules $\RRR = \{ r_1, \dots, r_m\}$, and let $\{\text{len}(r_1), \dots, \text{len}(r_m)\}$ be their respective length (that is, number of split tests), then the  \emph{complexity} of the explanation is computed as:
\begin{equation} \label{eq:avg-complexity}
    \CCC = \sum_{r_i \in \RRR} \text{len}(r_i)
\end{equation}

In practice, which set of rules $\RRR$ constitute an explanation is not always well defined and can depend on the user needs. In the context of our study, since we are focusing on local explanations, we include in $\RRR$ only the rules that are effectively used for the final instance prediction. More specifically:
\begin{itemize} \label{text:complexity}

    \item In case one or more decision trees are shown to the end-user as an explanation, we estimate $\CCC$ by counting only the rule(s) $r_{t_1}, \dots , r_{t_n}$ that lead to a leaf node, and we define $\RRR = \{ r_{t_1}, \dots r_{t_n} \}$. This is the case of HS, DT, C443, OOB Trees, and Small RF.
    
    \item In case, such as in RULECOSI+, the end-user is presented an ordered decision list with $m-1$ disjoint rules and a final rule $r_m$ that serves as an ``else" clause, then one of the two scenarios holds:
        \begin{itemize} \setlength{\itemsep}{0pt}
            \item the activated rule is ${r_n} \neq r_m$. In this case, we consider that the activated rule is a sufficient explanation, hence we set $\RRR = \{ r_n\}$;
            \item the activated rule is $r_m$, which is not a sufficient explanation since it has no antecedent and its prediction is used because none of the previous rules is activated. The explanation for the end-user is complete only if all rules are shown, therefore we set $\RRR = \{r_1, \dots r_m\}$.
        \end{itemize}
    \item In case of a collection of decision rules $\{ r_1, \dots, r_m\}$ such as in Bellatrex and SIRUS, we have $\RRR = \{ r_1, \dots, r_m\}$.
\end{itemize}

The above definition suggests that the complexity of an explanation increases as the rules get longer or as the cardinality of $\RRR$ increases, which is in agreement with human intuition. 
Regarding this, we notice that, let $d$ be the maximum depth of a tree and $K$ the number of extracted rules by Bellatrex for a given instance, we have $\CCC \leq K d$. This means that the complexity of the final explanation can be tuned by either adjusting the maximum depth of the underlying RF, or by tuning the maximum value of $K$ in the rule extraction step of Bellatrex. Such two-fold tuning can also be performed with C443, but cannot be as effective with other methods such as SIRUS (where rules are already extremely short), or DT and HS (where $K=1$ already).

\subsubsection*{Dissimilarity}

Finally, we consider the dissimilarity of the extracted rules. 
For this purpose, we define a distance on the tree-representation space. We use a generalised version of the Jaccard similarity index~\cite{Sies2020}, where the ratio between the element-wise minimum and maximum of two vectors is taken. Namely, given the vector representations $\bm{v_i}$ and $\bm{v_j}$ of two rules, we compute their similarity $\mathcal{S}$ as:
\begin{equation} \label{eq:Jaccard}
    \mathcal{S}(\bm{v_i},\bm{v_j}) = \frac{\sum_{k=1}^{p} \min(v_{i k}, v_{j k})}{\sum_{k=1}^{p} \max(v_{i k}, v_{j k})}.
\end{equation}
Given an instance $\bm{x}$ and the vector representation of the associated final rules $\bm{v}_{i_1}, \dots, \bm{v}_{i_K}$, we obtain the rule dissimilarity $\DDD$ for $\bm{x}$ by computing the average pairwise dissimilarity $1- \mathcal{S}$ between the $K$ final rule vector representations:
\begin{equation} \label{eq:avg-dissimilarity}
    \DDD = \frac{1}{K (K - 1)} \sum_{l \neq j}^{l, j \in \{ i_1, \dots i_K \}} 1-\mathcal{S}(\bm{v_l}, \bm{v_j})
\end{equation}
Finally, $\DDD$ is computed for every test instance, and the average is reported in the results (Section~\ref{sec:Results}). When $K=1$, dissimilarity in Equation~\eqref{eq:avg-dissimilarity} is not defined, therefore such instances are not included in the computation.




\subsection{Experimental set-up} \label{sec:Experiments-set-up}
In all experiments, we trained the original RF with 100 base learners.
As stopping criterion, we required that split-nodes included at least 5 instances, or at least 10 for the time-to-event scenario. No post-pruning techniques were applied.

The average predictive performance, complexity, as well as dissimilarity along a 5-fold cross validation is reported. The test sample size for datasets exceeding 500 instances is limited to 100 for computational reasons\footnote{Note that the algorithmic procedure of rule extraction, including hyperparameter tuning, is performed for every test instance separately.}. In the multi-label classification scenario, we drop labels that do not occur in the training or testing folds, since AUROC would otherwise not be defined.


With regards to hyperparameter tuning for the steps of our method (rule pre-selection, dimensionality reduction, clustering), the values are tested in a grid search fashion and are shown in Table~\ref{tab:hyperparameters}.
\begin{table}[ht]
\footnotesize
\centering
\begin{tabular}{|c|c|c|}
\hline
name & description & values \\
\hline
$\tau$ & \makecell{nb. of pre- \\ selected trees} & \{20, 50, 80\} \\
\hline
$d$ & \makecell{nb. of dimensions \\ after PCA} & \{2, 5, no PCA\}  \\
\hline
$K$ & \makecell{nb. of clusters \\ and final rules} & \{1, 2, 3\} \\
\hline
\end{tabular}
\caption{Possible choices for the hyperparameters in Bellatrex. For every instance the combination with the highest achieved fidelity $\mathcal{F}$ is chosen.}
\label{tab:hyperparameters}
\end{table}
As explained earlier, these values are optimised for each test instance individually according to fidelity (Equation~\ref{eq:fidelity}).

The rationale behind this grid is that the possible values for $\tau$ and $d$ cover a wide range while using at most 3 different values and limit computation time.
Furthermore, the values for $K$ are chosen so that at most 3 final rules are selected, keeping the model fairly explainable.

\section{Results}\label{sec:Results}

In this Section, we report the average predictive performance, complexity, and dissimilarity of the explanations generated by our method across the different datasets. On the predictive performance side, we compare the weighted and the unweighted path-based approach of Bellatrex against the competing methods introduced in Section~\ref{sec:Competing-methods}, with the RF model be treated as an upper, black-box benchmark.
Once the best performing rule representation approach for Bellatrex is identified, its explainability and dissimilarity are compared against relevant competitors.
More specifically, the complexity (Eq.~\eqref{eq:avg-complexity}) of the output explanations is compared against rule extracting algorithms where the total number of splits can be counted.
Finally, the average dissimilarity (Eq.~\eqref{eq:avg-dissimilarity}) of the rulesets extracted by Bellatrex is compared against methods that extract a similar number of rules from a RF.

Tables with dataset-specific results are presented for the binary classification data, whereas, to avoid excessive repetition, only average results are reported for the remaining scenarios. The reader can find the dataset-specific results in the Appendix. 

Additionally, we test the statistical significance of the observed differences (in performance, in dissimilarity and in complexity of the explanations) among methods, and we do so by conducting a post-hoc Friedman-Nemenyi test, setting the significance level to 0.05, as recommended in~\cite{Demsar2006}. 
Results are visualised with critical difference diagrams, which connect methods that are not statistically significantly different by horizontal line segments. Following the same approach as before, we report the results of the Friedman-Nemenyi for the binary classification tasks in the text, and we share the results of the remaining scenarios in the Appendix.

Finally, we present an ablation study; that is, we verify whether the main steps of the proposed procedure are of added value to the method. The study is repeated for the five tasks evaluated.


\subsection{Predictive performance} \label{sec:performances}

We compute the average test-set performance for every considered dataset over 5-fold cross validation. We report the average across datasets for each of the five types of prediction tasks, and we include the dataset-specific performance results for the binary classification classification case in Table~\ref{tab:binary-perf}.

\subsubsection*{Prediction performance in binary classification}


\begin{figure*}
    \scriptsize
    \centering

        \caption{Average predictive performance (AUROC) across binary classification datasets. Our proposed methods is shown under the name ``Bellatrex" in its two approaches. The highest achieved
        performances except for the ``black box" RF are shown in bold.}
        \label{tab:binary-perf}
        \begin{tabular}{l|cccccccccc}
        \toprule
         & RF & \textbf{\makecell{Bella- \\ trex}} &  \makecell{OOB \\ Trees} & \makecell{Single \\ DT} & \makecell{Small \\ RF} & LR & SIRUS & C443 & \makecell{Rule \\ COSI+} & HS \\
        \midrule
        blood & 0.7080 & 0.7041 & 0.6474 & 0.7048 & 0.6413 & \textbf{0.7394} & 0.7327 & 0.6256 & 0.6419 & 0.7209 \\
        breast cancer diagn. & 0.9836 & \textbf{0.9867} & 0.9393 & 0.9601 & 0.9526 & 0.9645 & 0.9806 & 0.9748 & 0.9400 & 0.9655 \\
        breast cancer & 0.9958 & 0.9954 & 0.9665 & 0.9626 & 0.9696 & 0.9970 & \textbf{0.9974} & 0.9923 & 0.9749 & 0.9625 \\
        breast cancer progn. & 0.5015 & 0.5348 & 0.5689 & 0.5893 & 0.5230 & \textbf{0.6370} & 0.5896 & 0.5504 & 0.5374 & 0.6333 \\
        breast cancer coimba & 0.7569 & 0.7508 & 0.6846 & 0.7108 & 0.6085 & \textbf{0.8108} & 0.7415 & 0.6762 & 0.5938 & 0.7200 \\
        Colon. Green & 0.9333 & \textbf{0.9449} & 0.8128 & 0.6962 & 0.7526 & 0.6718 & 0.7974 & 0.8333 & 0.7154 & 0.7897 \\
        Colon. Hinselm. & 0.5917 & 0.6229 & 0.5875 & 0.6146 & 0.4667 & 0.4833 & \textbf{0.6458} & 0.5813 & 0.5812 & 0.5792 \\
        Colon. Schiller & 0.6277 & 0.5923 & 0.5369 & 0.5600 & 0.5385 & 0.6615 & 0.6662 & \textbf{0.6969} & 0.6000 & 0.5246 \\
        divorce & 0.9941 & 0.9471 & 0.9412 & 0.9349 & 0.9405 & \textbf{1.0000} & 0.9920 & 0.9716 & 0.9325 & 0.9294 \\
        Flowmeters & 0.9943 & \textbf{0.9771} & 0.9271 & 0.9086 & 0.7800 & 0.4771 & 0.9286 & 0.9157 & 0.9514 & 0.9314 \\
        haberman & 0.6897 & 0.6786 & 0.6400 & 0.6185 & 0.6269 & 0.6856 & \textbf{0.6883} & 0.6763 & 0.5550 & 0.6175 \\
        hcc-survival & 0.8331 & \textbf{0.8250} & 0.6688 & 0.7162 & 0.7377 & 0.4192 & 0.7831 & 0.6369 & 0.6008 & 0.7138 \\
        ionosphere & 0.9828 & \textbf{0.9812} & 0.9204 & 0.9276 & 0.9067 & 0.9020 & 0.9406 & 0.9420 & 0.8251 & 0.9386 \\
        LSVT voice & 0.8824 & \textbf{0.8882} & 0.7404 & 0.7699 & 0.7228 & 0.6191 & 0.7993 & 0.7846 & 0.7022 & 0.7566 \\
        mamographic & 0.8523 & 0.8510 & 0.8247 & 0.8355 & 0.8249 & 0.8316 & 0.8439 & 0.8352 & 0.8072 & \textbf{0.8539} \\
        musk & 0.9511 & \textbf{0.9475} & 0.8300 & 0.7716 & 0.8088 & 0.9132 & 0.7562 & 0.8633 & 0.7158 & 0.7983 \\
        parkinson & 0.9349 & \textbf{0.9128} & 0.7912 & 0.7819 & 0.7766 & 0.3051 & 0.7424 & 0.8099 & 0.7663 & 0.7789 \\
        risk factors & 0.9709 & 0.9339 & 0.9050 & 0.9379 & 0.8745 & \textbf{0.9801} & 0.9420 & 0.7176 & 0.8179 & 0.9408 \\
        simulation crashes & 0.9284 & 0.9012 & 0.8335 & 0.7594 & 0.8027 & \textbf{0.9654} & 0.8663 & 0.5799 & 0.7939 & 0.7700 \\
        sonar & 0.9163 & \textbf{0.9117} & 0.7591 & 0.7498 & 0.7347 & 0.8455 & 0.8045 & 0.7493 & 0.7617 & 0.7622 \\
        SPECT & 0.7695 & 0.7489 & 0.7411 & 0.6957 & 0.7037 & \textbf{0.7870} & 0.7305 & 0.7682 & 0.7169 & 0.7502 \\
        SPECTF & 0.8203 & \textbf{0.7920} & 0.6193 & 0.6485 & 0.7273 & 0.7442 & 0.7838 & 0.7087 & 0.6656 & 0.6816 \\
        vertebral column & 0.9550 & 0.9477 & 0.9002 & 0.8907 & 0.8530 & \textbf{0.9562} & 0.8954 & 0.8423 & 0.8962 & 0.8762 \\
        wholesale & 0.9554 & \textbf{0.9525} & 0.9056 & 0.9322 & 0.9141 & 0.9023 & 0.9382 & 0.9364 & 0.9163 & 0.9390 \\
        \midrule 
        Average & 0.8554 & \textbf{0.8470} & 0.7788 & 0.7782 & 0.7578 & 0.7625 & 0.8161 & 0.7779 & 0.7504 & 0.7889 \\
        \bottomrule
        \end{tabular}
\end{figure*}

\begin{figure}[ht]
\centering
\includegraphics[width=0.8\columnwidth]{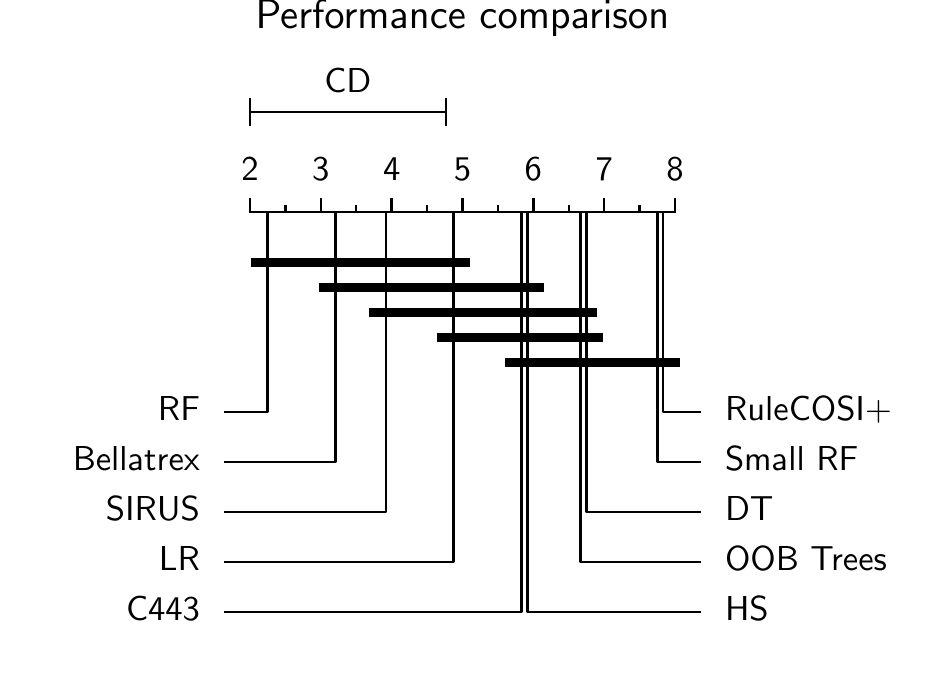}
\caption{Friedman-Nemenyi test, results regarding performance on binary classification data}
\label{fig:binary-FN-perform}
\end{figure}

We observe that Bellatrex considerably outperforms all competing methods and its performance is often on par, and sometimes better, than the original RF. In most of the cases, Bellatrex performs better than the other competing methods. Some exceptions to this trend are provided by the LR learner which, despite achieving non-remarkable performance on average, appears to be the best model in some datasets (e.g. ``blood" and ``Colon. Schiller"). This phenomenon might be explained by the intrinsically different (linear) nature of LR predictions as opposed to the other tree-based models.
In the middle range of the performance spectrum, we encounter competing methods from literature such as HS, with C443 and SIRUS scoring a couple of ``wins" and performing better than LR on average. On the lower end, we have the remaining methods: RuleCOSI+, Small RF, OOB Trees and DT. The trends from this qualitative analysis are confirmed by the Friedman-Nemenyi test $(\alpha = 0.05)$ and the resulting diagram is shown in Figure~\ref{fig:binary-FN-perform}.

We can observe that the difference between RF and Bellatrex is not statistically significant, meaning that we manage to provide interpretability without significant loss in predictive performance. It is also remarkable that Bellatrex is ranked on top of all the competitor methods and is also statistically significantly different to RuleCOSI+, Small RF, OOB Trees, as well as the single DT. 

This is to be expected as the competing methods offer a global approach for explanations, whereas Bellatrex follows a local approach. Furthermore, the Bellatrex instance-specific procedure is optimised towards increasing the fidelity to the original black-box model, which is a necessary condition for achieving interpretability~\cite{Guidotti2018}.

\subsubsection*{Prediction performance in TTE, regression, MLC, MTR}

Similar trends are highlighted when running Bellatrex on time-to event (TTE), regression, multi-label classification (MLC), and multi-target regression (MTR) data, with the average results being shown in Table~\ref{tab:avg-perf-all}. The competitiveness of Bellatrex against all competing methods is confirmed as Bellatrex outperforms all considered competitors. Furthermore, Bellatrex performance is on par compared to the black-box R(S)F in TTE tasks, furthermore Bellatrex is outperforming the original black-box in regression and MTR tasks; finally, a lower performance is observed for MLC tasks.

\begin{table}[ht]
    \centering
    \tiny
    \begin{tabular}{l|cccc}
    \multirow{2}{*}{Method} & \multicolumn{4}{c}{predictive scenario} \\
    & survival & multi-label & regression & multi-target \\
    \midrule
    \makecell{R(S)F} & 0.7274 & 0.8473 & 0.0679 & 0.0531 \\
    \midrule
    \makecell{Bellatrex \\ weighted} & \textbf{0.7275} & \textbf{0.7894} & 0.0674 & \textbf{0.0511} \\
    \midrule
    \makecell{Bellatrex \\ simple} & 0.7265 & 0.7888 & \textbf{0.0673} & 0.0516 \\
    \midrule
    \makecell{OOB (S) \\ Trees} & 0.6719 & 0.7524 &0.0772 & 0.0562 \\
    \midrule
    \makecell{Small \\ R(S)F} & 0.6724 & 0.7441 & 0.0788 & 0.0591 \\
    \midrule
    \makecell{Single \\ (S)DT} & 0.6696 & 0.7378 & 0.0832 & 0.0593 \\
    \midrule
    \makecell{Linear \\ Predictor\tablefootnote{Cox-PH for time-to-event tasks, LR for the other cases}} & 0.7040 & 0.5076 & 0.0942 & 0.0758 \\
    \midrule
    \makecell{SIRUS} & - & - & 0.1090 & - \\
    \midrule
    \makecell{HS} & - & 0.7585 & 0.0890 & 0.0715 \\
    \bottomrule
    \end{tabular}
    \footnotetext[1]{regularised Cox PH for survival scenario, regularised LR for the other scenarios}
    \caption{Average performance across remaining datasets. In bold, the best achieved performance excluding the black-box R(S)F}
    \label{tab:avg-perf-all}
\end{table}

This trend is to be expected, since methods like HS are based on a single DT and are therefore offering an extremely simplified prediction process, whereas methods based on LR fail to capture non-linearity and label correlations. Finally, SIRUS has been implemented for binary classification and regression problems, but has not been tested for the latter.
The results of the Friedman-Nemenyi statistical tests of these four scenarios are shown in Figure~\ref{fig:perf-nemenyi-all}.

Given that in Bellatrex the weighted path-based approach has a slightly better performance (and a more intuitive vector representation) compared to the simple one, we report from now on only the results of the former, unless otherwise specified.



\subsection{Complexity} \label{sec:complexities}


Next, we consider the complexity of explanations and compare Bellatrex against other competing methods. It is worth reminding that our definition of complexity in Eq.~\eqref{eq:avg-complexity} counts the number of split-rules performed across (possibly multiple) extracted rules. This allows to compare Bellatrex not only against tree-based methods such as DTs, Small RF and OOB Trees, but also against methods from literature that output rule-based explanation in a different format. More specifically, we can compare Bellatrex against SIRUS 
, HS 
, RuleCOSI+ 
, and C443 
. We do not report the complexity $\CCC$ from RF given its black-box nature and given that its complexity is much higher than the other reported methods.

\subsubsection*{Explanation complexity in binary classification}

Specifically for binary classification tasks, we compare Bellatrex against all aforementioned methods. To determine which rules are considered as part of an explanation, we refer to our guidelines in Section~\ref{sec:complexity}. We report average complexity $\CCC$  in Table~\ref{tab:binary-complexity}, and we also include the average number of rules extracted by Bellatrex.

\begin{figure*}[ht]
\centering
\scriptsize
\caption{Average complexity of the explanations across binary classification datasets.}\label{tab:binary-complexity}
\begin{tabular}{l|cccccccccc}
\toprule
 data & \makecell{Bellatrex \\ \tiny (rules)} &  \makecell{OOB \\ Trees} & \makecell{Small \\ RF} & DT & {\tiny SIRUS} & \makecell{\tiny Rule \\ \tiny COSI+} & C443 & HS \\
\midrule
blood & 15.13 (1.9) & 14.94 & 14.40 & 6.39 & 12.00 & 4.09 & 22.59 & \textbf{3.39} \\
breast cancer diagn. & 5.49 (1.3) & 5.79 & 5.93 & 4.13 & 13.40 & \textbf{3.48} & 13.10 & 4.05 \\
breast cancer  & 5.53 (1.2) & 6.13 & 5.00 & 4.01 & 11.60 & \textbf{3.45} & 13.48 & 4.85 \\
breast cancer progn. & 9.79 (2.0) & 10.56 & 11.19 & \textbf{4.13} & 10.00 & 9.13 & 15.61 & 4.33 \\
breast cancer coimba & 9.37 (2.2) & 9.57 & 9.36 & \textbf{3.39} & 11.20 & 8.17 & 13.89 & 4.32 \\
colonoscopy Green & 7.03 (1.9) & 8.44 & 8.23 & \textbf{3.76} & 10.20 & 5.98 & 13.07 & 4.82 \\
colonoscopy Hinselm. & 6.28 (1.7) & 6.94 & 6.31 & \textbf{2.71} & 10.20 & 4.23 & 10.92 & 4.28 \\
colonoscopy Schiller & 6.70 (1.9) & 7.97 & 7.08 & \textbf{3.03} & 10.20 & 4.83 & 11.08 & 3.41 \\
divorce & 1.47 (1.0) & 1.45 & 1.73 & \textbf{1.29} & 10.40 & 1.53 & 4.64 & 1.39 \\
Flowmeters & 4.89 (1.7) & 5.44 & 4.76 & \textbf{2.61} & 11.60 & 2.76 & 8.27 & 2.78 \\
haberman & 13.48 (1.9) & 12.88 & 13.12 & 5.73 & 13.20 & 7.25 & 19.71 & \textbf{5.22} \\
hcc-survival & 9.56 (2.2) & 10.87 & 11.00 & 4.56 & 10.00 & 7.32 & 15.64 & \textbf{4.32} \\
ionosphere & 8.01 (1.5) & 10.37 & 9.27 & \textbf{5.08} & 14.60 & 7.40 & 17.54 & 5.27 \\
LSVT voice & 6.60 (1.9) & 7.20 & 7.06 & \textbf{3.22} & 10.20 & 5.51 & 10.48 & 3.72 \\
mamographic & 16.21 (1.7) & 16.34 & 14.89 & 7.82 & 12.00 & 8.36 & 21.33 & \textbf{4.08} \\
musk & 12.48 (1.9) & 13.72 & 13.79 & 6.09 & 12.80 & 11.59 & 21.32 & \textbf{5.76} \\
parkinson & 13.39 (1.7) & 14.14 & 13.59 & 9.42 & 10.80 & 8.58 & 23.54 & \textbf{4.10} \\
risk factors & 7.27 (1.2) & 8.73 & 9.25 & 5.49 & 15.60 & \textbf{5.22} & 23.54 & 6.57 \\
simulation crashes & 6.48 (1.3) & 6.71 & 6.58 & \textbf{3.61} & 13.00 & 3.78 & 15.86 & 4.46 \\
sonar & 8.29 (2.0) & 9.33 & 9.62 & \textbf{3.74} & 10.40 & 9.42 & 14.04 & 4.02 \\
SPECT & 11.55 (1.8) & 12.04 & 12.04 & \textbf{4.17} & 15.00 & 5.40 & 11.87 & 4.77 \\
SPECTF & 7.85 (1.8) & 8.62 & 8.78 & \textbf{4.17} & 10.80 & 7.27 & 13.48 & 5.32 \\
 vertebral & 7.25 (1.6) & 7.42 & 7.46 & \textbf{3.70} & 11.80 & 6.17 & 14.20 & 4.37 \\
wholesale & 6.74 (1.4) & 6.65 & 6.31 & 4.35 & 13.00 & \textbf{2.96} & 14.87 & 4.52 \\
\midrule
average & 8.62 (1.7) & 9.26 & 9.03 & 4.44 & 11.83 & 6.00 & 15.17 & \textbf{4.34} \\
\bottomrule
\end{tabular}
\end{figure*}

The results show that Bellatrex needs $8.62$ splits on average to generate an explanation.
This finding, combined with the fact that $1.7$ rules are extracted on average suggests that the final rules are usually fairly short and can be considered more interpretable than a single rule of similar total complexity.
Furthermore, we also notice that the datasets for which Bellatrex extracts the least amount of rules (such as ``divorce", ``breast cancer original" and ``risk factors") are also the ones whose binary label is easy to predict (AUROC above $0.90$) for Bellatrex (cfr. Table~\ref{tab:binary-perf}). This is coherent with the principle of non-redundancy of extracted rules that Bellatrex seems to achieve: if a single rule is enough to make accurate predictions, there should be no advantage into extracting more rules.

\begin{figure}[ht]
    \centering
    \includegraphics[width=0.9\columnwidth]{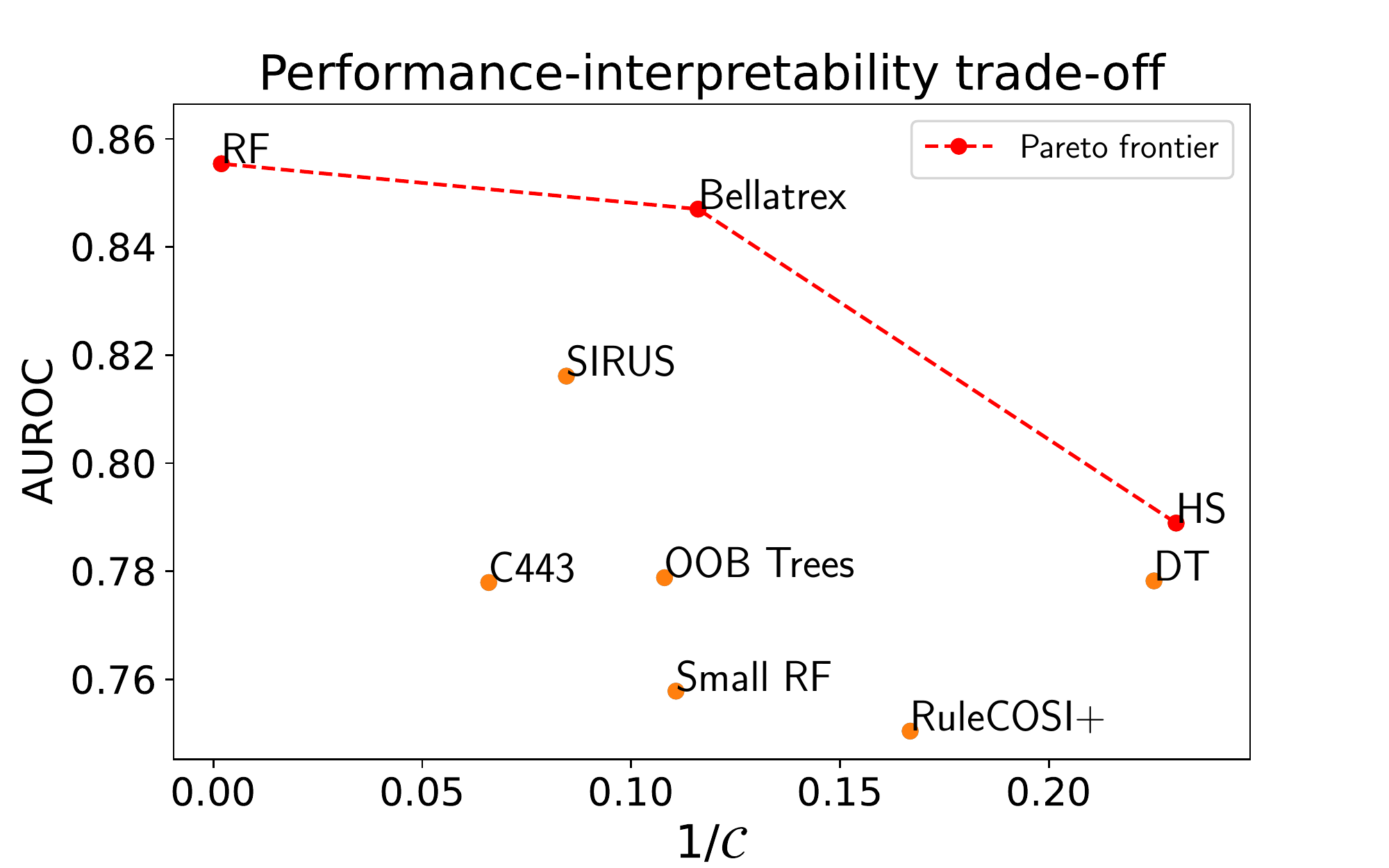}
    \caption{Visualisation of the performance vs. interpretability trade-off, where interpretability is approximated as $1/\CCC$. The performance-interpretability Pareto optimal frontier is shown in red.}
    \label{fig:trade-off-binary}
\end{figure}

When comparing Bellatrex with RuleCOSI+, C443, DT, and HS, we observe that the latter two generate shorter explanations. This is to be expected since these methods use a single rule to perform predictions for a given instance. It is worth noting however, that shorter explanations come at the expense of losing performance: we are therefore witnessing another instance of interpretability-performance trade-off (see Figure~\ref{fig:trade-off-binary}).
The aforementioned trade-off is also present when comparing the results of SIRUS against HS and RuleCOSI+: SIRUS outputs rules with higher total complexity, and at the same time shows a slightly higher performance. It is worth noting that SIRUS extracts 10 rules by default, and since its average number of rulesplits (complexity) is $11.83$, it means that the vast majority of the extracted rules has depth 1, consistent with what is reported in the original paper~\cite{Benard2021}. As for C443, the achieved performance is similar to HS, whereas its total complexity is comparable to SIRUS and higher than Bellatrex. The latter results is to be expected given that C443 always extracts 3 trees and therefore 3 rules are contributing to the computation of $\CCC$, as opposed to the average 1.69 rules in Bellatrex.

Finally, OOB Trees and Small RF achieve similar complexity compared to Bellatrex, which is to be expected given that the same number of rules is extracted from a similarly constructed underlying RF.
Such differences are not significant according toto the Friedman-Nemenyi statistical test,
whereas SIRUS generates statistically significantly more rule-splits compared to Bellatrex,
RuleCOSI+ and HS. The full results can be visualised in Figure~\ref{fig:complex-nemenyi-all}.

\subsubsection*{Explanation complexity in TTE, regression, MLC, MTR}

Continuing on the comparison of the (proxy for) interpretability, we compare the complexity of explanations $\CCC$ on the remaining four scenarios. We compare Bellatrex against HS for regression, multi-label, and multi-target regression scenarios, 
whereas SIRUS is, for the moment being, only suitable for use in (single target) regression tasks.
An overview of the comparison is given in Table~\ref{tab:avg-complexity-all}, whereas the full results can be found in Tables~\ref{tab:survival-complexity}-\ref{tab:mtr-complexity}.

\begin{table}[ht]
    \centering
    \tiny
    \begin{tabular}{l|cccc}
    \multirow{2}{*}{Method} & \multicolumn{4}{c}{predictive scenario} \\
    & survival & multi-label & regression & multi-target \\
    \midrule
    \makecell{Bellatrex \\ weighted} & 19.23 & 63.00 & 22.53 & 23.10 \\
    \midrule
    \makecell{OOB (S) \\ Trees} & 18.73 & 61.77 & 22.32 & 23.48 \\
    \midrule
    \makecell{Small \\ R(S)F} & 18.80 & 60.82 & 22.30 & 23.54 \\
    \midrule
    \makecell{Single \\ (S)DT} & 10.69 & 15.56 & 8.61 & 8.00 \\
    \midrule
    \makecell{SIRUS} & - & - & 13.53 & - \\
    \midrule
    \makecell{HS} & - & \textbf{5.14} & \textbf{4.27} & \textbf{4.15} \\
    \bottomrule
    \end{tabular}
    \caption{Average number of rule-splits used as explanation, comparison across datasets. In bold, the shortest average explanation per scenario.}
    \label{tab:avg-complexity-all}
\end{table}

The complexity of the rules extracted by Bellatrex and OOB (Survival) Trees and Small R(S)F is similar across all scenarios, in line with what is observed with the binary classification datasets. We also observe that Bellatrex generates on average longer explanations when run on multi-label datasets, and this can be due to the fact that multi-label datasets are more difficult to learn and need larger trees to fit. A confirmation of this can be found by inspecting the data: longest explanations are associated to larger datasets, with either a high number of instances or a high number of targets to predict (see Table~\ref{tab:multi-label-info}). For this case, we suggest the end user to select one (or a few) label of interest in the training phase, and  the Bellatrex explanations on the reduced label set will be shorter.

Finally, it is worth mentioning that the increase in average rule length observed in multi-label datasets is common to Bellatrex, DT, OOB Trees and Small RF, whereas it does not affect HS, whose rules are significantly shorter. Detailed results of the Friedman-Nemenyi test are shared in Figure~\ref{fig:complex-nemenyi-all}.

\subsection{Dissimilarity} \label{sec:dissimilarities}

Finally, we compare the average dissimilarity $\DDD$ of the final rules extracted by Bellatrex against the ones generated by Small RF, the ones picked according to the OOB Trees method, and the ones extracted by the C443 method.
These methods are indeed the only ones that extract a comparable number of independent learners; HS on the other hand builds a single decision tree whereas SIRUS extracts at least 10 rules. Finally, RULECOSI+ does not offer a meaningful comparison since the rules that it generates are not independent: exactly one of them can be activated, under the condition that the previous ones have not been used.


\subsubsection*{Rule dissimilarity in binary classification}

The detailed results for the binary classification scenario are shown in Table~\ref{tab:binary-dissim}.

\begin{table}[ht]
\centering
\tiny
    \begin{tabular}{l|cccc}
    \toprule
     &  \makecell{OOB \\ Trees} & \makecell{Small \\ RF} & Bellatrex & C443 \\
    \midrule
    blood & 0.5398 & 0.5743 & \textbf{0.6366} & 0.6080 \\
    B.C. diagnostic & 0.8799 & 0.8966 & 0.9569 & \textbf{0.9797} \\
    B.C. original & 0.7563 & 0.7349 & 0.8534 & \textbf{0.9027} \\
    B.C. prognostic & 0.9266 & 0.9251 & 0.9635 & \textbf{0.9696} \\
    B.C. coimba & 0.7862 & 0.7894 & 0.8769 & \textbf{0.8923} \\
    Col. Green & 0.9241 & 0.9561 & 0.9882 & \textbf{0.9955} \\
    Col. Hinselm. & 0.8747 & 0.9636 & 0.9739 & \textbf{0.9866} \\
    Col. Schiller & 0.9462 & 0.9659 & \textbf{0.9933} & 0.9892 \\
    divorce & 0.8936 & 0.9349 & \textbf{1.0000} & 0.9959 \\
    Flowmeters & 0.8513 & 0.9477 & 0.9901 & \textbf{1.0000} \\
    haberman & 0.5380 & 0.5686 & 0.6589 & \textbf{0.7616} \\
    hcc-survival & 0.9439 & 0.9381 & 0.9683 & \textbf{0.9998} \\
    ionosphere & 0.8948 & 0.9068 & 0.9450 & \textbf{0.9741} \\
    LSVT voice & 0.9826 & 0.9901 & \textbf{0.9971} & 0.9811 \\
    mamographic & 0.5175 & 0.5605 & 0.6185 & \textbf{0.6367} \\
    musk & 0.9773 & 0.9679 & 0.9894 & \textbf{0.9979} \\
    parkinson & 0.9867 & 0.9922 & 0.9963 & \textbf{0.9996} \\
    risk factors & 0.8167 & 0.8778 & 0.9414 & \textbf{0.9832} \\
    simul. crashes & 0.7532 & 0.8415 & 0.9400 & \textbf{0.9896} \\
    sonar & 0.9281 & 0.9555 & 0.9848 & \textbf{0.9983} \\
    SPECT & 0.8617 & 0.8656 & 0.8979 & \textbf{0.9389} \\
    SPECTF & 0.9231 & 0.9424 & 0.9608 & \textbf{0.9823} \\
    vertebral & 0.6531 & 0.6970 & 0.7868 & \textbf{0.8439} \\
    wholesale & 0.6871 & 0.7425 & 0.8277 & \textbf{0.8813} \\
    \midrule
    average & 0.8268 & 0.8556 & 0.9044 & \textbf{0.9286} \\
    \bottomrule
    \end{tabular}
    \caption{Average dissimilarity index across binary classification datasets. The dissimilarities achieved by our proposed methods are shown under the name ``Bellatrex".}
    \label{tab:binary-dissim}
\end{table}

When analysing average dissimilarity, we observe that C443 and Bellatrex achieve the highest dissimilarity, with C443 scoring most of the wins. These two methods consistently achieve higher dissimilarity compared to Small RF and OOB Trees. This means that
the Bellatrex algorithm is successful at picking dissimilar enough rulesets, and that these are not redundant.
We can verify how this dissimilarity in rules translates in practice in Section~\ref{sec:Examples}.

The Friedman-Nemenyi tests show that C443 and Bellatrex show a statistically significantly higher dissimilarity compared to OOB Trees and to Small RF, whose dissimilarities are driven by the intrinsic randomness of the RF algorithm. Detailed results are shared in Figure~\ref{fig:dissim-nemenyi-all}. Finally, the difference in dissimilarity between rulesets in C443 and rulesets in Bellatrex is not statistically significant.

\subsubsection*{Rule dissimilarity in TTE, regression, MLC, MTR}

The dissimilarity enforced on the final rules can be again compared against Small R(S)F and OOB (Survival) Trees in the other four types of prediction tasks, with the full results being available in the Appendix in Tables~\ref{tab:survival-dissimil}-\ref{tab:mtr-dissim}, and the respective post-hoc statistical tests in Figure~\ref{fig:dissim-nemenyi-all}.

\begin{table}[ht]
\tiny
\centering
\caption{Average dissimilarities comparison across datasets. In bold, the highest average dissimilarity per scenario}
\label{tab:avg-dissimil-all}
\begin{tabular}{l|cccc}
\multirow{2}{*}{\makecell{ \\ Method}} & \multicolumn{4}{c}{predictive scenario} \\
\cmidrule{2-5}
& survival & multi-label & regression & multi-target \\
\midrule
\makecell{Bellatrex \\ weighted} & \textbf{0.8420} & \textbf{0.9569} & \textbf{0.5459} & \textbf{0.7145} \\
\midrule
\makecell{OOB (S) \\ Trees} & 0.8219 & 0.9240 & 0.4171 & 0.5977 \\
\midrule
\makecell{Small \\ R(S)F} & 0.7945 & 0.9380 & 0.4483 & 0.6326 \\
\bottomrule
\end{tabular}
\end{table}

Bellatrex again outperforms Small R(S)F and OOB (Survival) Trees for every considered scenario. Running the Friedman-Nemenyi statistical tests shows that such differences are always significant except when considering the time-to-event datasets.

It is also worth noticing that high dimensional datasets generate the most dissimilar trees on average, and all datasets with $p > 100$ covariates are associated to average dissimilarity close to $1$, indicating almost no overlap in the splitting covariates across candidates. This trend is visible in all five considered scenarios, and is well illustrated in datasets like \textsf{parkison} ($\DDD = 0.9963$) for binary classification tasks, or \textsf{DLBCL} ($\DDD = 0.9994$) for time-to-event data.



\subsection{Ablation study} \label{sec:ablations}


As thoroughly described above, the Bellatrex method consists of four main steps: 1) pre-selection of the most relevant decision rules within the random forest, 2) mapping of the selected rules to a vector representation, 3) projection of the vector representations to a low-dimensional space and 3) the clustering step followed by the final rule extraction. The vector representation and the clustering steps are fundamental for our approach and cannot possibly be removed. To this end, in this ablation study we evaluate the added value of the rule pre-selection mechanism and of the dimensionality reduction step.

Hence, we have performed an ablation study where each step is separately employed measuring each time the end performance of our method. More specifically, we have compared four scenarios:
\begin{itemize}
    \item We apply our proposed method in full, including all steps described above;
    \item We apply dimensionality reduction but no pre-selection (i.e., no step~1);
    \item We apply the rule pre-selection step but no dimensionality reduction (i.e. no step~3);
    \item We apply neither dimensionality reduction (PCA) nor rule pre-selection (i.e., a baseline version of our method where both steps 1 and 3 are removed);
\end{itemize}


Here, we present the results related to binary classification datasets (Table \ref{tab:ablation-binary}) while the results related to the other tasks are provided in the Appendix (Tables~\ref{tab:ablation-survival}\,-\ref{tab:ablation-multitarget}). As shown, the best average performance is achieved when all steps of the proposed approach are included, affirming the added value of each of them. Furthermore, we observe the most substantial decrease in average performance when we remove both steps 1 and 3, whereas the intermediate scenarios (one step removed) fall somewhere in between.

\begin{table}[ht]
\centering
\tiny
\begin{tabular}{lrrrr}
\toprule
 & \makecell{\textbf{Bellatrex} \\ (original)} & \makecell{no pre\\ selection} & \makecell{no dim. \\ reduction} & \makecell{none of \\the two} \\
\midrule
blood & 0.7041 & \textbf{0.7050} & 0.7005 & 0.6975 \\
B.C. diagn. & 0.9867 & \textbf{0.9872} & 0.9865 & 0.9823 \\
B.C. original & \textbf{0.9954} & 0.9934 & 0.9953 & 0.9922 \\
B.C. progn. & 0.5348 & 0.5226 & \textbf{0.5441} & 0.4822 \\
B.C. coimba & 0.7508 & 0.7677 & \textbf{0.7723} & 0.7600 \\
Col. Green & \textbf{0.9449} & 0.8987 & 0.8385 & 0.9205 \\
Col. Hinselm. & 0.6229 & 0.4708 & 0.6042 & \textbf{0.6396} \\
Col. Schiller & 0.5923 & \textbf{0.6754} & 0.6077 & 0.6569 \\
divorce & \textbf{0.9471} & 0.9457 & 0.9412 & 0.9401 \\
Flowmeters & \textbf{0.9771} & 0.9514 & 0.9500 & 0.9286 \\
haberman & 0.6786 & 0.6726 & 0.6769 & \textbf{0.6832} \\
hcc-survival & \textbf{0.8250} & 0.7638 & 0.8058 & 0.7412 \\
ionosphere & \textbf{0.9812} & 0.9768 & 0.9737 & 0.9444 \\
LSVT voice & \textbf{0.8882} & 0.8654 & 0.8243 & 0.8618 \\
mamographic & \textbf{0.8510} & 0.8467 & 0.8494 & 0.8396 \\
musk & \textbf{0.9475} & 0.9298 & 0.9230 & 0.8757 \\
parkinson & \textbf{0.9128} & 0.8824 & 0.9005 & 0.8340 \\
risk factors & 0.9339 & 0.9309 & \textbf{0.9351} & 0.9339 \\
simul. crashes & \textbf{0.9012} & 0.8875 & 0.8629 & 0.7775 \\
sonar & \textbf{0.9117} & 0.8928 & 0.8864 & 0.8141 \\
SPECT & 0.7489 & 0.7524 & 0.7500 & \textbf{0.7680} \\
SPECTF & 0.7920 & \textbf{0.8177} & 0.7918 & 0.7478 \\
vertebral & 0.9477 & \textbf{0.9501} & 0.9499 & 0.9414 \\
wholesale & 0.9525 & \textbf{0.9549} & 0.9529 & 0.9455 \\
\midrule
average & \textbf{0.8470} & 0.8351 & 0.8343 & 0.8212 \\
\bottomrule
\end{tabular}
\caption{Ablation study results on the binary datasets. In bold, the best achieved performance.}
\label{tab:ablation-binary}
\end{table}

\section{Illustration of Bellatrex} \label{sec:Examples}

In this Section, we show how Bellatrex works in practice, and to do so we pick two examples from the ``boston housing" regression dataset~\cite{BostonDataset}. This dataset contains $506$ neighbourhoods (``towns"), each described by 14 covariates. The target variable is the median house value, normalised to the $[0,1]$ interval.

\begin{figure}[ht]
    \centering
    \includegraphics[width=0.95\columnwidth]{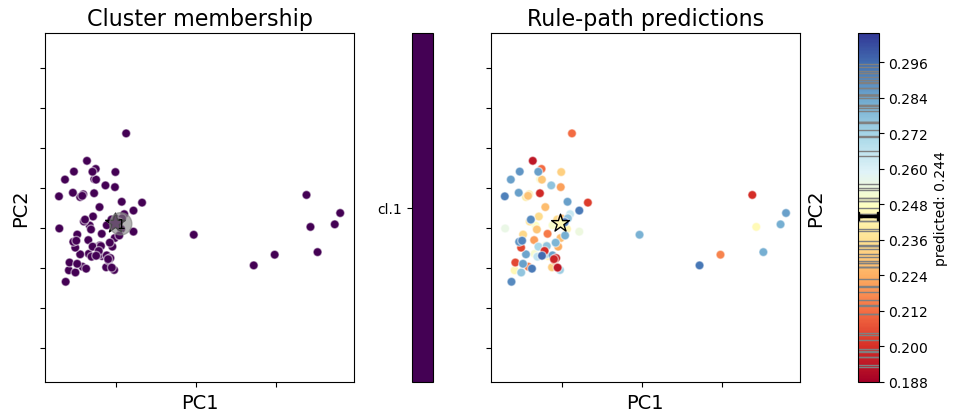}
    \caption{Bellatrex performed on a test instance of the \textsf{boston housing} dataset. On the left, the grey circle corresponds to the centre of the cluster; on the right, rules are coloured according to their predicted value. On both sides, the final extracted rule, closest to the centre of the cluster, is shown in a star shape.}
    \label{fig:boston-example41}
\end{figure}

Our first example consists of a test instance with a relatively low house value of $0.206$, for which Bellatrex predicts a value of $0.244$. 
More specifically, Bellatrex reaches the prediction by discarding 20 rules in the pre-selection step and projecting the vector representation of the remaining 80 rules in 2 dimensions; such representations are grouped in a single cluster and the rule closest to the centre is selected as the final representative.
Further insights are shown in Figure~\ref{fig:boston-example41}, where the rule representations are shown.

The left plot of the Figure shows that a single cluster is selected, and that the four rules on the right side of the plot are  being treated as outliers. 
On the right-side plot, the same rules are coloured according to their prediction, and these are shown to vary between $0.20$ and $0.29$.
The final representative rule is situated in the central position and is associated to a prediction close to the average value of $0.24$. Inspection of such rule is shown in Example~\ref{tab:boston-example-path}, where we show how, starting from an initial estimated equal to the average (bootstrapped) value of the root node, moves until a final value indicated by the corresponding leaf.

\begin{example}[ht]
    \centering
    \tiny
    \begin{tabular}{lcl}
    rule & (weight=1) & \makecell[l]{initial \\ estimate $=0.390$} \\
    \midrule
    Instance & split test & prediction\\
    \midrule
    (n. rooms = 5.64) & n. rooms $\leq 6.80$ & $\rightarrow 0.317$ \\
    (low status = 18.34) & low status $> 14.30$ & $\rightarrow 0.209$ \\
    (low status = 18.34) & low status $\leq 19.08$ & $\rightarrow 0.271$ \\
    (age = 94.70) & age $> 93.90$ & $\rightarrow 0.238$ \\
    (crime rate = 0.88) & crime rate $\leq 2.15$ & $\rightarrow 0.275$ \\
    (low status = 18.34) & low status $> 17.07$ & $\rightarrow$ \textbf{0.244} (leaf) \\
    \bottomrule
    \end{tabular}
    \caption{Extracted rule.}
    \label{tab:boston-example-path}
\end{example}

The extracted rule is aligned with human intuition: the root node-split shows that a smaller average number of rooms (\textsf{n. rooms}) per dwelling is associated to a lower house value, with the prediction dropping from an initial estimate of $0.390$ to  $0.317$. Similarly, the splits involving \textsf{low status} show that houses situated in a poorer neighbourhood (that is, with high values of \textsf{low status}), are associated to lower prices. According to this rule, the median house price of the town increases with a low \textsf{crime rate}, and decreases with high proportion of houses being built before 1940 (high value for the \textsf{age} variable).

Our second example involves a test instance of high value $y=0.660$, whose Bellatrex prediction is $0.625$ (and underlying RF prediction is $0.633$), and two rules are selected as the final prediction, as shown in Figure~\ref{fig:boston-example45} below.
\begin{figure}[ht]
    \centering
    \includegraphics[width=0.95\columnwidth]{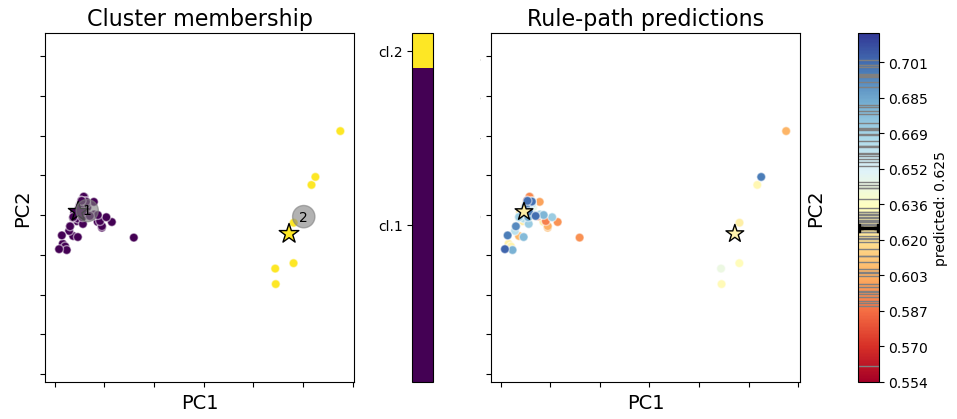}
    \caption{Bellatrex performed on a test instance of the \textsf{boston housing} dataset. The final extracted rules are shown with a star shape.}
    \label{fig:boston-example45}
\end{figure}

The plot shows that two final rules are extracted, one from the large cluster on the left side, and one for the small cluster on the right side. In both cases, the predictions are close to the average, and in general,
we obtain similar predictions from rules following different splits and structures.
The extracted rules are shown in Example~\ref{tab:boston-example-path2}.

The first rule follows human intuition when assigning higher house value for higher  \textsf{n. rooms} and for lower \textsf{crim. rate} rates. Furthermore, it also reflects the racial bias of the 1970 US Census~\cite{BostonDataControversy} when assigning higher value to ethnically homogeneous neighbourhoods (high values of \textsf{black} covariate).

\begin{example}
    \begin{center}
    \tiny
    \begin{tabular}{lcl}
    rule 1 & ($w_1=0.90$) & \makecell[l]{initial \\ estimate $=0.399$} \\
    \midrule
    Instance & split test & prediction\\
    \midrule
    (n. rooms = 7.18) & n. rooms $> 6.98$ & $\rightarrow 0.774$ \\
    (n. rooms = 7.18) & n. rooms $\leq 7.44$ & $\rightarrow 0.647$ \\
    (black = 393) & black $> 386$  & $\rightarrow 0.6642$ \\
    (n. rooms = 7.18) & n. rooms $\leq 7.26$ & $\rightarrow 0.679$ \\
    (crime rate = 0.03) & crime rate $\leq 0.03$ & $\rightarrow \textbf{0.625}$ (leaf)\\
    \end{tabular}
    \begin{tabular}{lcl}
    \toprule
    rule 2& ($w_2=0.10$) & \makecell[l]{initial \\ estimate $=0.379$} \\
    \midrule
    Instance & split test & prediction\\
    \midrule
    (low status = 4.03) & low status $\leq 9.71$ & $\rightarrow 0.549$ \\
    (n. rooms = 7.18) & n. rooms $\leq 7.44$ & $\rightarrow 0.500$ \\
    (n. rooms = 7.18) & n. rooms $> 6.73$ & $\rightarrow 0.603$ \\
    (pupils ratio = 17.8) & pupils ratio $< 18.9$ & $\rightarrow 0.624$ \\
    (age = 61.10) & age $\leq 87.00$ & $\rightarrow 0.612$ \\
    (black = 393) & black $> 392$  & $\rightarrow 0.641$ \\
    (age = 61.10) & age $> 41.95$ & $\rightarrow 0.608$ \\
    (n. rooms = 7.18) & n. rooms $> 6.93$ & $\rightarrow 0.630$ \\
    (pupils ratio = 17.8) & pupils ratio $> 15.50$ & $\rightarrow \textbf{0.628}$ (leaf) \\
    \bottomrule
    \end{tabular}
    \end{center}
    \caption{Extracted rules.}
    \label{tab:boston-example-path2}
\end{example}

The second rule shows a similar perspective by using a different set of tests: the trend highlighted by \textsf{n. rooms} is consistent with previous findings, whereas conditions including \textsf{low status}, \textsf{pupils ratio}, and \textsf{age} appear for the first time in this example. Consistently with human intuition lower values of pupils-to-teacher ratio (\textsf{pupils ratio}) are associated with higher house value, whereas the \textsf{age} value for this instance has a small negative effect on house pricing. Finally, consistently with the previous example, a lower value for \textsf{low status} is associated to higher median house value.

\section{Conclusions and future work} \label{sec:Conclusion}

In this article, we proposed a novel method (Bellatrex) that interprets random forest predictions by extracting, for a given instance, a surrogate model consisting of the most representative decision trees in the forest.
Although we have focused on the random forest ensemble model, our proposed approach is able to accommodate other tree-ensemble models such as Extremely Randomized Trees~\cite{ExtraTrees}. Furthermore, Bellatrex is compatible with other variations of random forests such as Random Survival Forest and Random Forest Regression and has been evaluated across 89 datasets belonging to five different predictive tasks: binary classification, time-to-event prediction, multi-label classification, regression, and multi-target regression. To our knowledge, no other related work is as versatile as Bellatrex in providing explanations in so many different scenarios, or at least, no such experimental results have been reported.

Bellatrex achieves high levels of predictive performance by extracting a small number of rules, which we limited to three in our experiments. 
Its performance is often on par with that of random forests, and even outperforms this benchmark in two out of five scenarios. 
We can therefore conclude that our method achieves its intended goal of explaining the prediction of a random forest with a few rules, without giving up on predictive performance. 


Furthermore, when compared against ante-hoc explainable methods that do not rely on a tree structure, such as LR and Cox-PH, Bellatrex comes on top.
Finally, the comparisons with state-of-the-art methods outline that Bellatrex outperforms, and often significantly, all considered methods from literature that aim at extracting explanations from a trained RF, namely C443~\cite{Sies2020}, HS~\cite{HS2022} and SIRUS~\cite{Benard2021}.

The analysis of the explanation complexity shows that Bellatrex extracts on average 2 rules with a total combined complexity (i.e. test-splits used) that is normally below 25. 
Such numbers suggest that our extracted rules are fairly short on average, 
even more so since two rules of length $\ell$ are more interpretable than a single rule of length $2\ell$. Rule post-pruning can further reduce total complexity and increase interpretability.
The comparison against other methods such as HS and RuleCOSI+ shows that the latter two generate shorter rules as explanations
while performing significantly worse (performance-interpretability trade-off). Finally, the SIRUS method generates explanations of comparable complexity to Bellatrex, but consists in many extremely short rules that are not as convenient to inspect.

We also studied the dissimilarity of the trees (or paths) extracted by Bellatrex.
We showed that Bellatrex has (often significantly) greater dissimilarity compared to picking the same number of random trees from the ensemble, or picking the same number of best performing trees.
This dissimilarity in Bellatrex is remarkable, as it proves to be successful despite the fact that the optimal number of clusters $K$ is not optimised for increasing dissimilarity, but rather for increasing fidelity to the original random forest. Furthermore, the high average dissimilarity suggests that the final extracted rules are not redundant, and that they give diverse possible explanations to the end user. 


Our method also has some limitations, which are similar to other methods that extract learners from a tree ensemble. Firstly, the interpretability of the rules extracted by Bellatrex is challenged when such rules are too long, and to tackle this limitation, the end-user can set more stringent stopping criteria to the RF. Alternatively, in case of multi-label tasks, the end user can select a subset of labels of interest in the training phase and the resulting RF will lead to shorter rules.
Another possible limitation lies in the computational complexity of the method: although a single prediction can be explained in a short time, running the procedure on a full dataset can become computationally expensive.


Directions for future work include exploring new vector representations,
as well as evaluating Bellatrex on an even broader set of tasks,
such as multi-event survival analysis~\cite{Multi-event-SA}, online learning~\cite{river}, or network inference \cite{Biclustering}.
Post-pruning techniques can be applied on top of the extracted rules
 of Bellatrex as a further improvement step, thanks to which we expect an
 increase of interpretability with shorter rules, and little to no price
 paid for predictive performance.


\appendix
\section{Appendix}

In the Appendix we include the results of the performance and dissimilarity results as well as the ablation studies of the remaining scenarios.


\subsection{Performance}

Here, we report the dataset-specific performance, as well as the outcomes of the statistical testing procedure, for time-to-event, multi-label classification, single target regression, and multi-target regression data.  Detailed results on binary classification data are reported in the main body, Section~\ref{sec:performances}.

\begin{sidewaystable}
\begin{center}
\tiny
\begin{minipage}{\textheight}
\caption{Average predictive performance (C-index) across time-to-event datasets. Our proposed methods is shown in under the name ``Bellatrex" in its two configurations: tree and path-based.}\label{tab:survival-perf}
\begin{tabular*}{\textheight}{@{\extracolsep{\fill}}l|ccccccccccc@{\extracolsep{\fill}}}
\toprule
dataset & RSF & \makecell{\textbf{Bellatrex} \\ \textbf{weighted}} & \makecell{\textbf{Bellatrex} \\ \textbf{simple}} & 
\makecell{OOB \\ S. Trees} & \makecell{Small \\ RSF} & \makecell{Single \\ SDT} & \makecell{Cox-PH} \\
\midrule
addicts & 0.6497 & 0.6512 & 0.6484 & 0.6065 & 0.6323 & 0.6205 & \textbf{0.6563} \\
B. C. survival & 0.6524 & 0.6417 & \textbf{0.6497} & 0.5619 & 0.5863 & 0.5619 & 0.5708 \\
DBCD & 0.7549 & 0.7534 & \textbf{0.7592} & 0.5940 & 0.5574 & 0.6686 & 0.7235 \\
DLBCL & 0.6352 & \textbf{0.6357} & 0.6185 & 0.5697 & 0.5796 & 0.5456 & 0.6129 \\
echocardiogram & 0.4145 & 0.4166 & 0.4181 & 0.4427 & \textbf{0.4958} & 0.4785 & 0.4096 \\
FLChain-single event & 0.8312 & \textbf{0.8330} & 0.8321 & 0.8105 & 0.7683 & 0.7858 & 0.8218 \\
gbsg2 & 0.7022 & \textbf{0.7035} & 0.7032 & 0.6850 & 0.6434 & 0.6419 & 0.6876 \\
lung & 0.6216 & \textbf{0.6236} & 0.6189 & 0.5816 & 0.5904 & 0.5386 & 0.6056 \\
NHANES I & 0.8205 & \textbf{0.8209} & 0.8193 & 0.7358 & 0.7435 & 0.7584 & 0.8043 \\
PBC & 0.8469 & 0.8467 & \textbf{0.8472} & 0.7977 & 0.7601 & 0.7780 & 0.8192 \\
rotterdam (excl. \textsf{recurr}) & 0.7954 & \textbf{0.7961} & \textbf{0.7961} & 0.7581 & 0.7518 & 0.7731 & 0.7668 \\
rotterdam (incl. \textsf{recurr}) & 0.9038 & \textbf{0.9047} & 0.9036 & 0.8774 & 0.8418 & 0.8652 & 0.8996 \\
veteran & 0.7349 & 0.7349 & 0.7352 & 0.6547 & 0.6983 & 0.6546 & \textbf{0.7452} \\
whas500 & 0.7449 & \textbf{0.7458} & 0.7444 & 0.6922 & 0.6730 & 0.6763 & 0.7332 \\
\midrule
average & \textbf{0.7224} & \textbf{0.7224} & 0.7214 & 0.6693 & 0.6680 & 0.6661 & 0.7044 \\
\bottomrule
\end{tabular*}
\end{minipage}
\end{center}
\end{sidewaystable}

\begin{sidewaystable}
\tiny
\begin{center}
\begin{minipage}{\textheight}
\caption{Average predictive performance (MAE)  
across regression datasets.}\label{tab:regress-perf}
\begin{tabular*}{\textheight}{@{\extracolsep{\fill}}l|ccccccccccc@{\extracolsep{\fill}}}
\toprule
dataset & RF & \makecell{\textbf{Bellatrex} \\ \textbf{weighted}} & \makecell{\textbf{Bellatrex} \\ \textbf{simple}} &  OOB Trees & Single DT & \makecell{Small \\ RF} & Ridge & SIRUS & HS \\
\midrule
airfoil & 0.0373 & 0.0374 & \textbf{0.0373} & 0.0492 & 0.0645 & 0.0493 & 0.1069 & 0.1195 & 0.0896 \\
AmesHousing & 0.0229 & 0.0230 & 0.0230 & 0.0280 & 0.0308 & 0.0297 & \textbf{0.0226} & 0.0504 & 0.0356 \\
auto mpg & 0.0525 & 0.0525 & \textbf{0.0523} & 0.0615 & 0.0644 & 0.0664 & 0.0676 & 0.0871 & 0.0692 \\
bike sharing & 0.0481 & \textbf{0.0481} & 0.0483 & 0.0586 & 0.0635 & 0.0594 & 0.0783 & 0.0933 & 0.0713 \\
boston housing & 0.0577 & \textbf{0.0576} & 0.0580 & 0.0650 & 0.0747 & 0.0724 & 0.0824 & 0.0842 & 0.0694 \\
california & 0.0613 & \textbf{0.0611} & 0.0614 & 0.0726 & 0.0848 & 0.0791 & 0.1040 & 0.1438 & 0.1045 \\
car imports & 0.0362 & \textbf{0.0360} & 0.0362 & 0.0436 & 0.0456 & 0.0417 & 0.0384 & 0.0666 & 0.0465 \\
Computer& 0.0291 & 0.0297 & \textbf{0.0289} & 0.0327 & 0.0423 & 0.0356 & 0.0354 & 0.0440 & 0.0369 \\
compressive strength & 0.0397 & \textbf{0.0396} & 0.0397 & 0.0497 & 0.0602 & 0.0534 & 0.1053 & 0.1126 & 0.0807 \\
concrete slump & 0.0735 & 0.0734 & 0.0731 & 0.0816 & 0.0852 & \textbf{0.0669} & 0.0903 & 0.1223 & 0.0890 \\
ENB2012 cooling & 0.0339 & \textbf{0.0338} & \textbf{0.0338} & 0.0350 & 0.0386 & 0.0363 & 0.0724 & 0.0768 & 0.0434 \\
ENB2012 heating & 0.0112 & \textbf{0.0111} & \textbf{0.0111} & 0.0114 & 0.0132 & 0.0127 & 0.0644 & 0.0807 & 0.0247 \\
forest fires & 0.3040 & 0.3032 & \textbf{0.3028} & 0.3100 & 0.3256 & 0.3248 & 0.3190 & 0.3170 & 0.3116 \\
PRSA data & 0.0359 & 0.0361 & \textbf{0.0358} & 0.0460 & 0.0491 & 0.0463 & 0.0882 & 0.1022 & 0.0855 \\
slump dataset & 0.0682 & 0.0680 & 0.0682 & 0.0778 & 0.0797 & 0.0901 & \textbf{0.0506} & 0.0957 & 0.0744 \\
students final math & 0.1556 & 0.1550 & \textbf{0.1546} & 0.1855 & 0.1851 & 0.1666 & 0.1764 & 0.1611 & 0.1624 \\
wine quality all & 0.0741 & \textbf{0.0710} & 0.0715 & 0.0879 & 0.0893 & 0.0871 & 0.0930 & 0.1025 & 0.0961 \\
wine quality red & 0.0843 & 0.0808 & \textbf{0.0807} & 0.0934 & 0.1008 & 0.0978 & 0.1029 & 0.1123 & 0.1067 \\
wine quality white & 0.0645 & \textbf{0.0623} & \textbf{0.0623} & 0.0774 & 0.0833 & 0.0810 & 0.0923 & 0.0996 & 0.0938 \\
\midrule
average & 0.0679 & 0.0674 & \textbf{0.0673} & 0.0772 & 0.0832 & 0.0788 & 0.0942 & 0.1090 & 0.0890 \\
\bottomrule
\end{tabular*}
\end{minipage}
\end{center}
\end{sidewaystable}

\begin{sidewaystable}
\tiny
\begin{center}
\begin{minipage}{\textheight}
\caption{Average predictive performance (AUROC) across multi-label datasets. Our proposed methods is shown in under the name ``Bellatrex" in its two configurations: tree and path-based.}\label{tab:multi-label-perf}
\begin{tabular*}{\textheight}{@{\extracolsep{\fill}}l|ccccccccccc@{\extracolsep{\fill}}}
\toprule
 & RF & \makecell{\textbf{Bellatrex} \\ \textbf{weighted}} & \makecell{\textbf{Bellatrex} \\ \textbf{simple}} &  OOB Trees & Single DT & Small RF & LR & HS \\
\midrule
birds & 0.9199 & 0.8190 & \textbf{0.8278} & 0.7862 & 0.7371 & 0.7818 & 0.4563 & 0.7565 \\
CAL500 & 0.5781 & \textbf{0.5428} & 0.5372 & 0.5327 & 0.5206 & 0.5349 & 0.5085 & 0.5427 \\
emotions & 0.8520 & 0.8205 & \textbf{0.8267} & 0.7586 & 0.7453 & 0.7498 & 0.4399 & 0.7846 \\
enron & 0.8192 & \textbf{0.7707} & 0.7566 & 0.7322 & 0.7190 & 0.7164 & 0.5071 & 0.7349 \\
flags & 0.7458 & 0.7237 & \textbf{0.7325} & 0.6882 & 0.6799 & 0.6336 & 0.6372 & 0.6938 \\
genbase & 1.0000 & \textbf{0.9977} & \textbf{0.9977} & 0.9928 & 0.9971 & 0.9944 & 0.5019 & 0.9976 \\
langlog & 0.7575 & 0.6075 & 0.5903 & 0.6012 & 0.6107 & 0.5991 & 0.5099 & \textbf{0.6940} \\
medical & 0.9841 & \textbf{0.9497} & 0.9409 & 0.9102 & 0.9279 & 0.8954 & 0.5182 & 0.9461 \\
ng20 & 0.9631 & 0.9225 & \textbf{0.9239} & 0.8525 & 0.8360 & 0.8630 & 0.4996 & 0.8180 \\
scene & 0.9477 & \textbf{0.9195} & 0.9165 & 0.8437 & 0.7865 & 0.8451 & 0.5023 & 0.8353 \\
slashdot & 0.8691 & 0.8186 & \textbf{0.8235} & 0.7669 & 0.7716 & 0.7754 & 0.4978 & 0.7263 \\
stackex chess & 0.8330 & 0.6928 & \textbf{0.6948} & 0.6670 & 0.6538 & 0.6653 & 0.5016 & 0.6916 \\
yeast & 0.7455 & 0.6777 & \textbf{0.6856} & 0.6491 & 0.6063 & 0.6197 & 0.5180 & 0.6393 \\
\midrule
average & 0.8473 & \textbf{0.7894} & 0.7888 & 0.7524 & 0.7378 & 0.7441 & 0.5076 & 0.7585 \\
\bottomrule
\end{tabular*}
\end{minipage}
\end{center}
\end{sidewaystable}

\begin{sidewaystable}
\tiny
\begin{center}
\begin{minipage}{\textheight}
\caption{Average predictive performance (weighted MAE) across multi-target datasets. Our proposed ``Bellatrex" is shown in both its weighted and unweighted approach.}\label{tab:mtr-perf}
\begin{tabular*}{\textheight}{@{\extracolsep{\fill}}l|ccccccccccc@{\extracolsep{\fill}}}
\toprule
 & RF & \makecell{\textbf{Bellatrex} \\ \textbf{weighted}} & \makecell{\textbf{Bellatrex} \\ \textbf{simple}} &  OOB Trees & Single DT & Small RF & Ridge & HS \\
\midrule
andro & 0.0959 & \textbf{0.0911} & 0.0933 & 0.1078 & 0.1162 & 0.1102 & 0.1695 & 0.1374 \\
atp1d & 0.0484 & \textbf{0.0479} & 0.0483 & 0.0556 & 0.0589 & 0.0563 & 0.0574 & 0.0648 \\
atp7d & 0.0470 & \textbf{0.0443} & 0.0454 & 0.0489 & 0.0503 & 0.0597 & 0.0711 & 0.0580 \\
edm & 0.1194 & \textbf{0.1068} & 0.1102 & 0.1072 & 0.1189 & 0.1259 & 0.1732 & 0.1323 \\
enb & 0.0227 & \textbf{0.0225} & 0.0227 & 0.0228 & 0.0244 & 0.0241 & 0.0641 & 0.0335 \\
ENB2012 & 0.0235 & \textbf{0.0234} & 0.0235 & 0.0249 & 0.0271 & 0.0258 & 0.0684 & 0.0353 \\
jura & 0.0678 & \textbf{0.0678} & 0.0681 & 0.0783 & 0.0808 & 0.0801 & 0.0727 & 0.0808 \\
oes10 & 0.0218 & 0.0220 & 0.0219 & 0.0246 & 0.0268 & 0.0263 & \textbf{0.0187} & 0.0294 \\
oes97 & 0.0293 & 0.0296 & 0.0298 & 0.0321 & 0.0343 & 0.0339 & \textbf{0.0262} & 0.0342 \\
osales & 0.0378 & \textbf{0.0361} & 0.0362 & 0.0422 & 0.0431 & 0.0433 & 0.0585 & 0.0461 \\
rf1 & 0.0023 & 0.0019 & \textbf{0.0018} & 0.0027 & 0.0036 & 0.0030 & 0.0324 & 0.0454 \\
rf2 & 0.0038 & \textbf{0.0031} & 0.0032 & 0.0051 & 0.0063 & 0.0053 & 0.0284 & 0.0444 \\
scm1d & 0.0253 & 0.0249 & \textbf{0.0247} & 0.0323 & 0.0362 & 0.0330 & 0.0365 & 0.0640 \\
scm20d & 0.0309 & \textbf{0.0296} & 0.0298 & 0.0402 & 0.0419 & 0.0388 & 0.0664 & 0.0817 \\
scpf & 0.0113 & \textbf{0.0105} & 0.0108 & 0.0138 & 0.0113 & 0.0163 & 0.0120 & 0.0135 \\
sf1 & 0.0772 & \textbf{0.0726} & 0.0736 & 0.0797 & 0.0727 & 0.0769 & 0.0749 & 0.0768 \\
sf2 & 0.0321 & 0.0318 & 0.0318 & 0.0324 & \textbf{0.0312} & 0.0328 & 0.0322 & 0.0339 \\
slump & 0.1451 & 0.1433 & 0.1438 & \textbf{0.1403} & 0.1643 & 0.1501 & 0.1588 & 0.1649 \\
wq & 0.1681 & \textbf{0.1610} & \textbf{0.1610} & 0.1774 & 0.1791 & 0.1806 & 0.1838 & 0.1828 \\
\midrule
average & 0.0531 & \textbf{0.0511} & 0.0516 & 0.0562 & 0.0593 & 0.0591 & 0.0740 & 0.0715 \\
\bottomrule
\end{tabular*}
\end{minipage}
\end{center}
\end{sidewaystable}

\begin{figure*}[ht]
    \centering
    \begin{subfigure}[b]{0.47\textwidth}
        \centering
        \includegraphics[width=\textwidth]{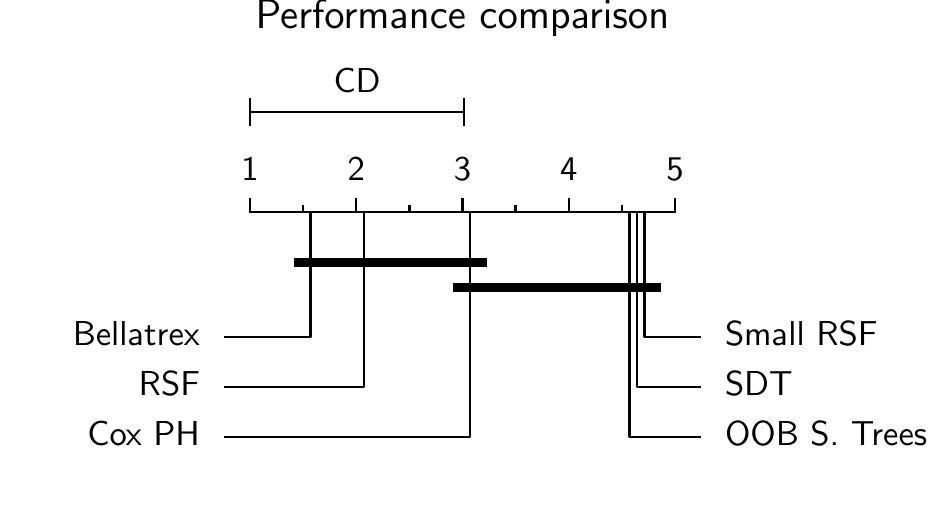}
        {{\small Nemenyi-test for time-to-event data}}    
        \label{fig:perf-nemenyi-surv}
    \end{subfigure}
    \hfill
    \begin{subfigure}[b]{0.47\textwidth}  
        \centering 
        \includegraphics[width=\textwidth]{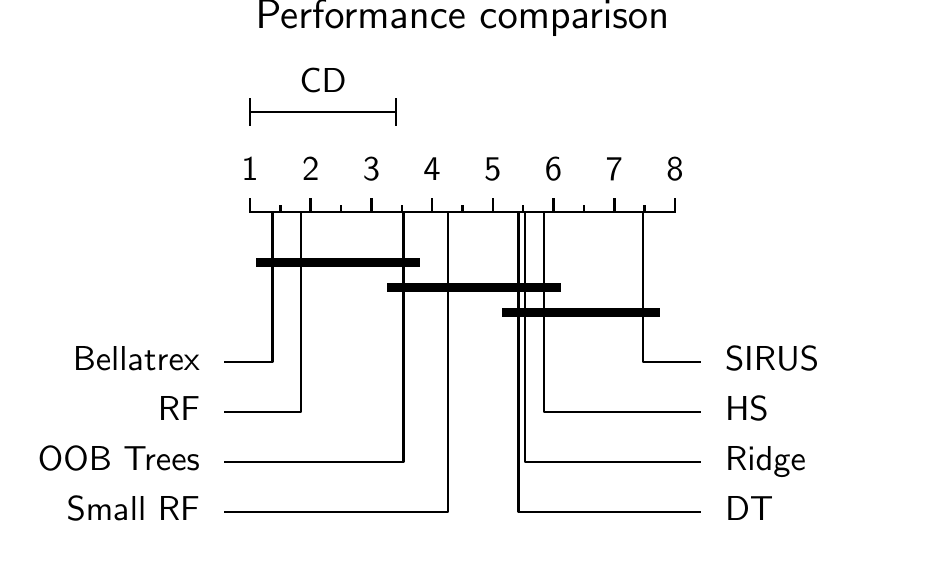}
        {{\small Nemenyi-test for regression tasks}}    
        \label{fig:perf-nemenyi-regress}
    \end{subfigure}
    \vskip\baselineskip
    \begin{subfigure}[b]{0.47\textwidth}   
        \centering 
        \includegraphics[width=\textwidth]{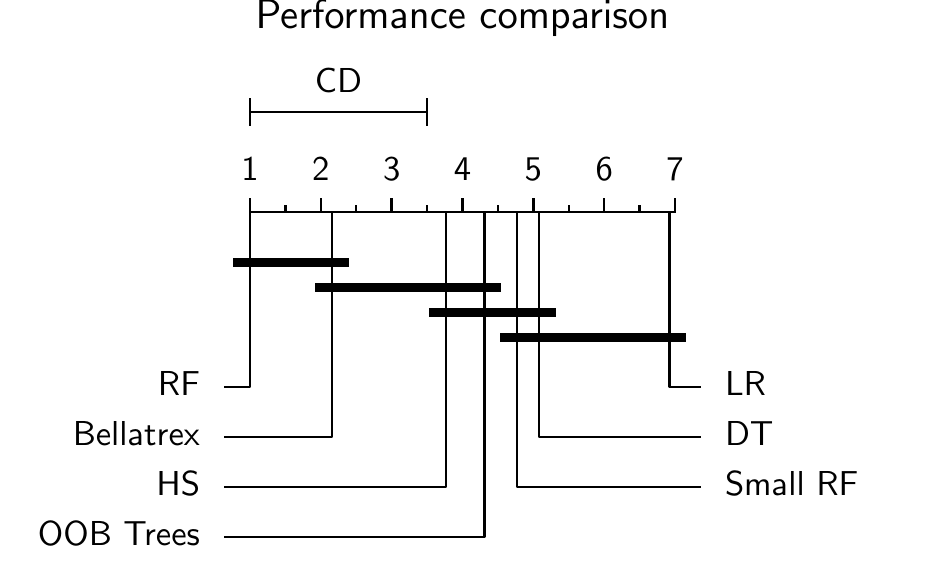}
        {{\small Nemenyi-test for multi-label classification data}}    
        \label{fig:perf-nemenyi-multi}
    \end{subfigure}
    \hfill
    \begin{subfigure}[b]{0.47\textwidth}   
        \centering 
        \includegraphics[width=\textwidth]{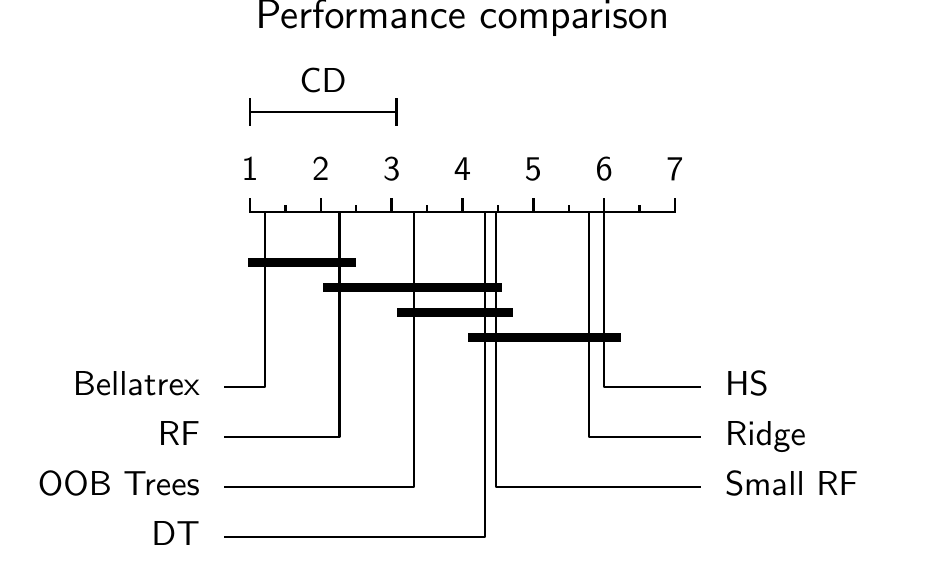}
        {{\small Nemenyi-test for multi-target regression data}}    
        \label{fig:perf-nemenyi-mtr}
    \end{subfigure}
    \vspace{0.4cm}
    \caption[fig:perf-nemenyi-all]
    {Friedman-Nemenyi statistical test, results on performance} 
    \label{fig:perf-nemenyi-all}
\end{figure*}

\FloatBarrier
\newpage


\subsection{Complexity}

Here, we report the dataset-specific results with regards to complexity of the explanations (cfr. Equation~\eqref{eq:avg-complexity}), as well as the outcomes of the statistical testing procedure, for time-to-event, multi-label classification, single target regression, and multi-target regression data.  Detailed results on binary classification data are reported in the main body, Section~\ref{sec:complexities}

\begin{table}[ht]
\centering
\tiny
\caption{Average complexity of explanations in time-to event data}\label{tab:survival-complexity}
\begin{tabular}{l|rrr}
\toprule
 & \textbf{\makecell{Bella- \\ trex }}  & \makecell{OOB \\ STrees} & \makecell{Small \\ RSF} \\
\midrule
addicts & 10.51 & \textbf{9.46} & 10.07 \\
B. C. survival & \textbf{10.82} & 11.02 & 11.73 \\
DBCD & 21.11 & \textbf{16.93} & 19.64 \\
DLBCL & 26.98 & 23.61 & \textbf{23.53} \\
echocardiogram & 10.15 & 10.18 & \textbf{9.73} \\
FLChain & 31.73 & 32.15 & \textbf{31.69} \\
gbsg2 & 19.04 & \textbf{17.79} & 18.01 \\
lung & 12.86 & \textbf{11.88} & 12.30 \\
NHANES I & 29.85 & 29.47 & \textbf{28.94} \\
PBC & \textbf{14.70} & 15.07 & 14.91 \\
rotterdam (excl. \textsf{recurr}) & 21.88 & 21.83 & \textbf{21.79} \\
rotterdam (incl. \textsf{recurr}) & 21.83 & 22.90 & \textbf{21.48} \\
veteran & \textbf{9.90} & 10.80 & 11.11 \\
whas500 & \textbf{15.76} & 15.89 & 15.79 \\
\midrule
average & 18.42 & \textbf{17.85} & 17.97 \\
\bottomrule
\end{tabular}
\end{table}

\begin{table}[ht]
\tiny
\begin{center}
\caption{Average complexity of explanations in regression datasets}
\label{tab:regress-complexity}
\begin{tabular}{l|rrrrr}
\toprule

 & \textbf{\makecell{Bella- \\ trex }}  & \makecell{OOB \\ Trees} & \makecell{Small \\ RF} & SIRUS &  HS \\
\midrule
airfoil & 23.10 & 22.62 & 22.46 & 13.80 & \textbf{4.17}  \\
AmesHousing & 29.85 & 29.11 & 28.85 & 14.00 & \textbf{3.85}  \\
auto mpg & 16.75 & 16.44 & 16.78 & 12.00 & \textbf{3.73}  \\
bike sharing & 21.86 & 21.57 & 21.49 & 14.80 & \textbf{4.46}  \\
boston housing & 21.33 & 21.77 & 20.57 & 12.20 & \textbf{3.58}  \\
california & 38.60 & 37.74 & 37.82 & 13.00 & \textbf{3.92}  \\
car imports & 14.93 & 15.69 & 15.77 & 14.20 & \textbf{4.76}  \\
Computer& 16.83 & 17.67 & 17.61 & 14.80 & \textbf{4.19}  \\
compress. strength & 22.77 & 22.25 & 22.79 & 12.20 & \textbf{4.13}  \\
concrete slump & 11.76 & 11.32 & 11.68 & 13.40 & \textbf{4.63}  \\
ENB2012 cooling & 17.69 & 17.70 & 17.64 & 14.20 & \textbf{3.92}  \\
ENB2012 heating & 17.32 & 17.34 & 17.25 & 15.00 & \textbf{4.36}  \\
forest fires & 20.31 & 20.40 & 20.24 & 10.00 & \textbf{5.88}  \\
PRSA data & 41.26 & 41.30 & 41.30 & 15.60 & \textbf{4.20}  \\
slump dataset & 12.27 & 12.07 & 12.55 & 13.40 & \textbf{4.86}  \\
students final math & 19.34 & 19.11 & 18.58 & 14.80 & \textbf{4.54}  \\
wine quality all & 29.28 & 28.45 & 28.09 & 14.20 & \textbf{3.71}  \\
wine quality red & 23.70 & 22.85 & 23.32 & 11.20 & \textbf{4.08}  \\
wine quality white & 29.23 & 28.77 & 28.82 & 14.20 & \textbf{4.20}  \\
\midrule
average & 22.54 & 22.32 & 22.30 & 13.53 & \textbf{4.27}  \\
\bottomrule
\end{tabular}
\end{center}
\end{table}

\begin{table}[ht]
\centering
\tiny
\caption{Average complexity of explanations in multi-label data.}\label{tab:multi-label-complexity}
\begin{tabular}{l|rrrrr}
\toprule
 & \textbf{ Bellatrex} &  \makecell{OOB \\ Trees} & \makecell{Small\\ RF} & HS \\
\midrule
birds & 23.43 & 23.97 & 24.72 & \textbf{4.82}  \\
CAL500 & 29.29 & 31.29 & 29.87 & \textbf{4.26}  \\
emotions & 19.48 & 21.84 & 22.61 & \textbf{4.40}  \\
enron & 57.84 & 59.84 & 56.64 & \textbf{4.72}  \\
flags & 16.51 & 17.65 & 18.10 & \textbf{4.27}  \\
genbase & 22.64 & 20.48 & 21.68 & \textbf{5.79}  \\
langlog & 42.98 & 39.50 & 38.19 & \textbf{7.02}  \\
medical & 59.49 & 56.71 & 56.08 & \textbf{6.06}  \\
ng20 & 127.97 & 122.78 & 119.65 & \textbf{6.30}  \\
scene & 25.95 & 26.71 & 26.88 & \textbf{3.38}  \\
slashdot & 248.33 & 244.98 & 240.03 & \textbf{5.50}  \\
stackex chess & 111.52 & 100.02 & 100.56 & \textbf{5.14}  \\
yeast & 33.51 & 37.18 & 35.60 & \textbf{5.10}  \\
\midrule
average & 63.00 & 61.77 & 60.82 & \textbf{5.14} \\
\bottomrule
\end{tabular}
\end{table}

\begin{table}[ht]
\centering
\tiny
\caption{Average complexity of explanations in multi-target data.}\label{tab:mtr-complexity}
\begin{tabular}{l|rrrr}
\toprule
 & \textbf{Bellatrex} & \makecell{OOB \\ Trees} & \makecell{Small\\ RF} & HS \\
\midrule
andro & 8.77 & 9.62 & 9.18 & \textbf{5.42}  \\
atp1d & 19.01 & 19.86 & 19.41 & \textbf{3.65}  \\
atp7d & 18.97 & 20.05 & 20.24 & \textbf{3.65}  \\
edm & 9.25 & 9.27 & 9.30 & \textbf{3.72}  \\
enb & 18.55 & 18.58 & 18.65 & \textbf{3.92}  \\
ENB2012 & 18.32 & 18.33 & 18.44 & \textbf{4.13}  \\
jura & 20.87 & 20.71 & 20.53 & \textbf{4.44}  \\
oes10 & 24.49 & 24.38 & 24.40 & \textbf{4.40}  \\
oes97 & 23.26 & 23.73 & 23.05 & \textbf{4.25}  \\
osales & 49.87 & 55.07 & 57.53 & \textbf{4.72}  \\
rf1 & 30.40 & 30.94 & 30.67 & \textbf{3.43}  \\
rf2 & 27.48 & 27.98 & 27.98 & \textbf{3.98}  \\
scm1d & 33.82 & 33.65 & 33.46 & \textbf{3.77}  \\
scm20d & 34.84 & 34.84 & 35.14 & \textbf{4.01}  \\
scpf & 27.75 & 27.13 & 27.71 & \textbf{3.37}  \\
sf1 & 12.42 & 12.34 & 12.88 & \textbf{4.55}  \\
sf2 & 17.05 & 17.11 & 16.44 & \textbf{4.55}  \\
slump & 13.72 & 13.21 & 13.30 & \textbf{4.89}  \\
wq & 30.08 & 29.28 & 29.04 & \textbf{4.09}  \\
\midrule
average & 23.10 & 23.48 & 23.54 & \textbf{4.15}  \\
\bottomrule
\end{tabular}

\end{table}

\begin{figure*}[ht]
    \centering
    \begin{subfigure}[t]{0.47\textwidth}
        \centering
        \includegraphics[width=\textwidth]{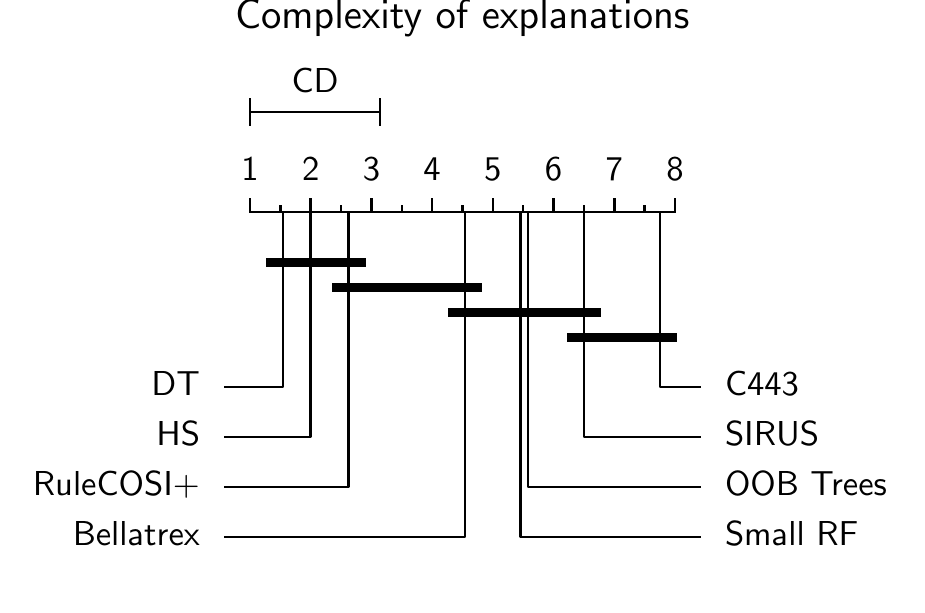}
        {{\small binary data}}    
        \label{fig:complex-nemenyi-bin}
    \end{subfigure}
    \vfill
    \vspace{0.2cm}
    \begin{subfigure}[b]{0.47\textwidth}
        \centering
        \includegraphics[width=\textwidth]{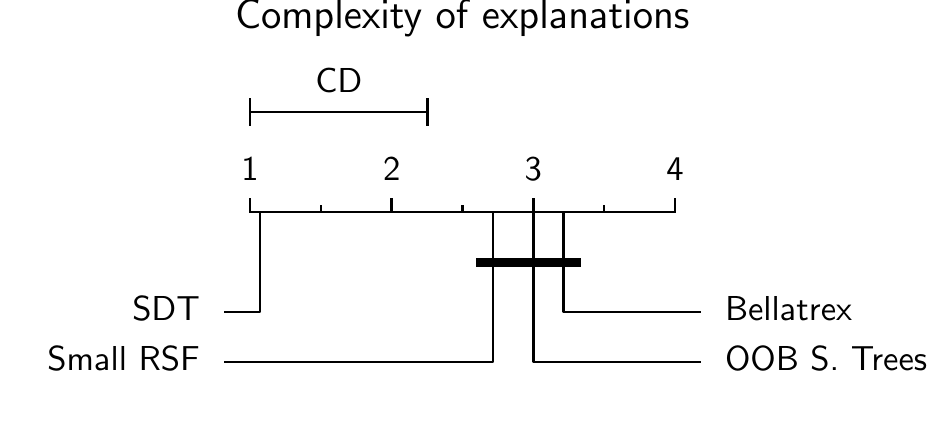}
        \captionsetup{width=.9\linewidth}
        {{\small time-to-event data}}
        \label{fig:complex-nemenyi-surv}
    \end{subfigure}
    \hfill
    \begin{subfigure}[b]{0.47\textwidth}  
        \centering 
        \includegraphics[width=\textwidth]{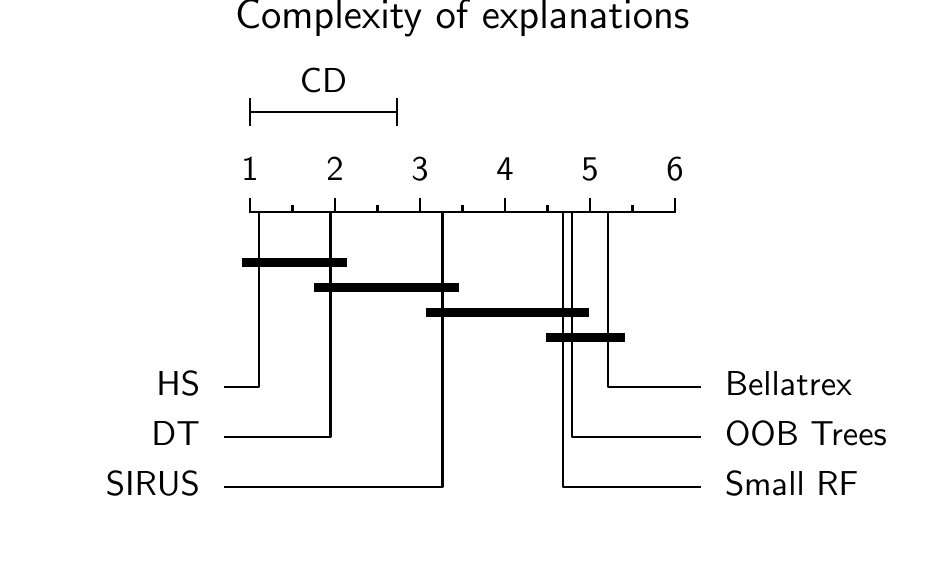}
        {{\small regression tasks}}    
        \label{fig:complex-nemenyi-regress}
    \end{subfigure}
    \vskip\baselineskip
    \vspace{0.4cm}
    \begin{subfigure}[b]{0.47\textwidth}   
        \centering 
        \includegraphics[width=\textwidth]{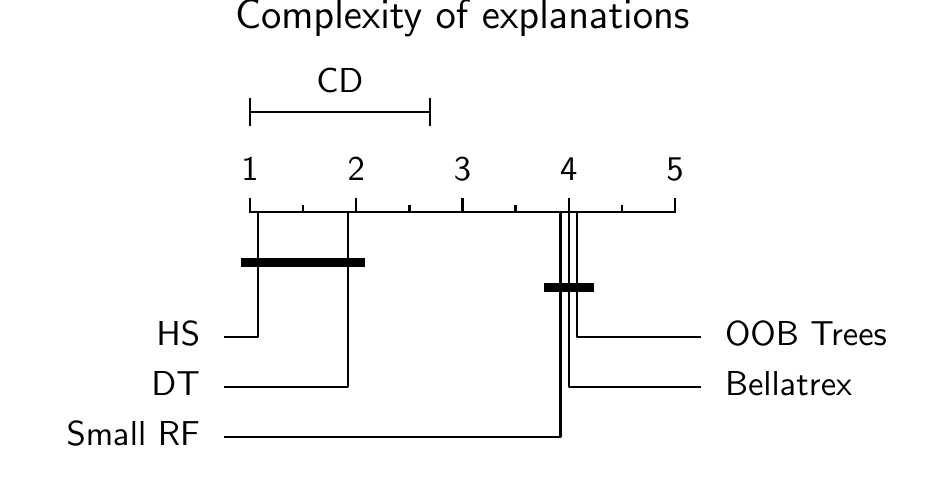}
        {{\small multi-label classification data}}    
        \label{fig:complex-nemenyi-multi}
    \end{subfigure}
    \hfill
    \begin{subfigure}[b]{0.47\textwidth}   
        \centering 
        \includegraphics[width=\textwidth]{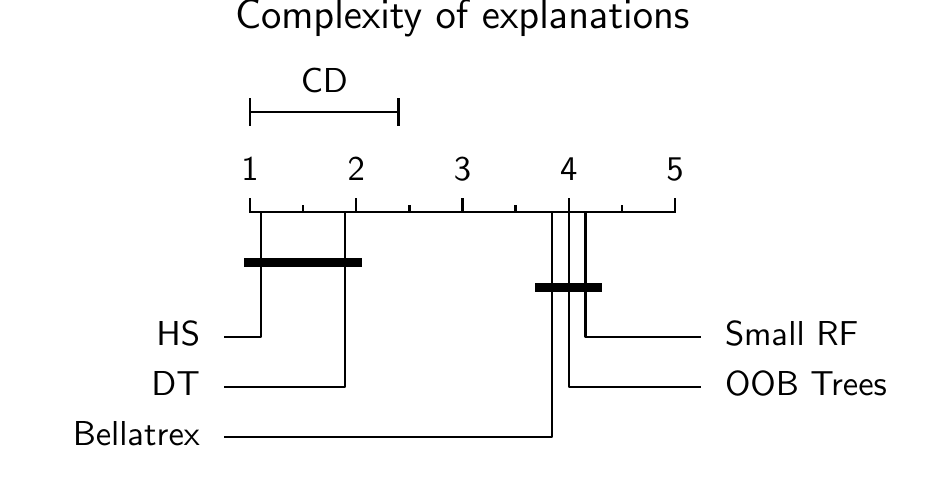}
        {{\small multi-target regression data}}
        \label{fig:complex-nemenyi-mtr}
    \end{subfigure}
    \vspace{0.4cm}
    \caption[fig:complex-nemenyi-all]
    {Nemenyi statistical test, performed on the complexity of the explanations across the five scenarios.}
    \label{fig:complex-nemenyi-all}
\end{figure*}

\FloatBarrier
\newpage

\subsection{Dissimilarity}

Here, we report the dataset-specific results results for dissimilarity (cfr. Equation~\eqref{eq:avg-dissimilarity}), as well as the outcomes of the statistical testing procedure, for time-to-event, multi-label classification, single target regression, and multi-target regression data.  Detailed results on binary classification data are reported in the main body, Section~\ref{sec:dissimilarities}.

\begin{table}[ht]
\centering
\tiny
\captionsetup{width=.9\linewidth}
\caption{Dissimilarities for time-to event data}\label{tab:survival-dissimil}
\begin{tabular}{l|ccc|c}
\toprule
 & \makecell{OOB \\ Trees} & \makecell{Small \\ RSF} & \textbf{\makecell{Bella- \\ trex }}  & \makecell{ average \\ n. rules} \\
\midrule
addicts & \textbf{0.6032} & 0.4419 & 0.5679 & 2.17 \\
B. C. survival & 0.9657 & 0.9563 & \textbf{0.9841} & 2.26 \\
DBCD & 0.9969 & 0.9978 & \textbf{0.9990} & 2.26 \\
DLBCL & 0.9876 & 0.9971 & \textbf{0.9994} & 2.40 \\
echocardiogram & 0.7952 & 0.7905 & \textbf{0.8567} & 2.25\\
FLChain & 0.6214 & 0.6761 & \textbf{0.7430} & 2.31\\
gbsg2 & \textbf{0.7691} & 0.7242 & 0.7633 & 2.33\\
lung & 0.8182 & 0.7743 & \textbf{0.8395} & 2.29\\
NHANES I & 0.7953 & 0.7945 & \textbf{0.8391} & 2.08 \\
PBC & 0.8495 & 0.8239 & \textbf{0.8664} & 2.28 \\
rotterdam (excl. \textsf{recurr})& \textbf{0.7958} & 0.7333 & 0.7614 & 2.18 \\
rotterdam (incl. \textsf{recurr}) & 0.7501 & 0.7541 & \textbf{0.8187} & 2.22 \\
veteran & 0.7904 & 0.7461 & \textbf{0.8167} & 2.33 \\
whas500 & \textbf{0.8503} & 0.7977 & 0.8305 & 2.35 \\
\midrule
average & 0.8135 & 0.7863 & \textbf{0.8347} & 2.27\\
\bottomrule
\end{tabular}
\end{table}

\begin{table}[ht]
    \tiny
\centering
\captionsetup{width=.9\linewidth}

\caption{Average rule dissimilarity for regression datasets}
\begin{tabular}{l|ccc|c}
\toprule
 &  \makecell{OOB \\ Trees} & \makecell{Small\\ RF} & \textbf{\makecell{Bella- \\ trex }}  & \makecell{average \\ n. rules} \\
\midrule
airfoil & 0.2391 & 0.2499 & \textbf{0.3236} & 2.30 \\
AmesHousing & 0.3034 & 0.3868 & \textbf{0.4994} & 2.35 \\
auto mpg & 0.6067 & 0.5759 & \textbf{0.7403} & 2.28 \\
bike sharing & 0.3403 & 0.4536 & \textbf{0.6360} & 2.31 \\
boston housing & 0.3728 & 0.4165 & \textbf{0.5223} & 2.34 \\
california & 0.1293 & 0.1833 & \textbf{0.2721} & 2.29 \\
car imports & 0.5419 & 0.4940 & \textbf{0.6270} & 2.31 \\
Computer & 0.4917 & 0.5141 & \textbf{0.5566} & 2.30 \\
compress. strength & 0.2432 & 0.3342 & \textbf{0.4340} & 2.38 \\
concrete slump & 0.3353 & 0.3921 & \textbf{0.3989} & 2.30 \\
ENB2012 cooling & 0.6035 & 0.6663 & \textbf{0.7583} & 2.31 \\
ENB2012 heating & 0.6049 & 0.6002 & \textbf{0.7552} & 2.26 \\
forest fires & 0.7254 & 0.7555 & \textbf{0.7967} & 2.27 \\
PRSA data & 0.2322 & 0.2378 & \textbf{0.4673} & 2.43 \\
slump dataset & 0.4581 & 0.5381 & \textbf{0.6753} & 2.41 \\
students final math & 0.6451 & 0.6821 & \textbf{0.8208} & 2.30 \\
wine quality all & 0.2955 & 0.2886 & \textbf{0.3160} & 2.02 \\
wine quality red & 0.4333 & 0.4422 & \textbf{0.4544} & 2.09 \\
wine quality white & \textbf{0.3232} & 0.3061 & 0.3185 & 2.11 \\
\midrule
average & 0.4171 & 0.4483 & \textbf{0.5459} & 2.28 \\
\bottomrule
\end{tabular}
\label{tab:regress-dissim}
\end{table}

\begin{table}[ht]
\centering
\tiny
\captionsetup{width=.9\linewidth}
\caption{Average dissimilarity of extracted rules in multi-label data.}\label{tab:multi-label-dissim}
\begin{tabular}{l|ccc|c}
\toprule
 &  \makecell{OOB \\ Trees} & \makecell{Small\\ RF} & \textbf{Bellatrex} & \makecell{average \\ n. rules} \\
\midrule
birds & 0.9544 & 0.9664 & \textbf{0.9859} & 2.19 \\
CAL5 & 0.9589 & 0.9494 & \textbf{0.9731} & 2.95 \\
emotions & 0.9318 & 0.9430 & \textbf{0.9772} & 2.63 \\
enron & 0.9652 & 0.9754 & \textbf{0.9847} & 2.73 \\
flags & 0.8615 & 0.8496 & \textbf{0.9115} & 2.76 \\
genbase & 0.9255 & 0.9364 & \textbf{0.9465} & 1.33 \\
langlog & 0.9690 & 0.9800 & \textbf{0.9886} & 2.40 \\
medical & 0.9507 & 0.9571 & \textbf{0.9700} & 2.57 \\
ng20 & 0.8614 & 0.9093 & \textbf{0.9169} & 2.54 \\
scene & 0.9543 & 0.9768 & \textbf{0.9908} & 2.38 \\
slashdot & 0.8606 & 0.8830 & \textbf{0.8873} & 2.52 \\
stackex chess & 0.8862 & 0.9272 & \textbf{0.9404} & 2.78 \\
yeast & 0.9330 & 0.9406 & \textbf{0.9662} & 2.78 \\
\midrule
average & 0.9240 & 0.9380 & \textbf{0.9569} & 2.51 \\
\bottomrule
\end{tabular}
\end{table}

\begin{table}[ht]
\centering
\tiny
\caption{Average rule dissimilarity of multi-target data.}\label{tab:mtr-dissim}
\begin{tabular}{l|ccc|c}
\toprule
 &  \makecell{OOB \\ Trees} & \makecell{Small\\ RF} & \textbf{Bellatrex} & \makecell{average \\ n. rules} \\
\midrule
andro & 0.7265 & 0.8548 & \textbf{0.9324} & 2.58 \\
atp1d & 0.8369 & 0.9082 & \textbf{0.9483} & 2.61 \\
atp7d & 0.8345 & 0.9477 & \textbf{0.9731} & 2.51 \\
edm & 0.5925 & 0.7311 & \textbf{0.8482} & 2.02 \\
enb & 0.5129 & 0.6408 & \textbf{0.7467} & 2.42 \\
ENB2012 & 0.6717 & 0.6377 & \textbf{0.7579} & 2.40 \\
jura & 0.4259 & 0.4799 & \textbf{0.5551} & 2.57 \\
oes10 & 0.9386 & 0.9285 & \textbf{0.9506} & 2.82 \\
oes97 & 0.9387 & 0.9394 & \textbf{0.9574} & 2.82 \\
osales & 0.6833 & 0.7192 & \textbf{0.7262} & 2.66 \\
rf1 & 0.1251 & 0.1270 & \textbf{0.1683} & 2.48 \\
rf2 & 0.1772 & 0.1918 & \textbf{0.2928} & 2.45 \\
scm1d & 0.5864 & 0.6428 & \textbf{0.8560} & 2.55 \\
scm20d & 0.5261 & 0.4319 & \textbf{0.6534} & 2.64 \\
scpf & 0.4680 & 0.4429 & \textbf{0.4682} & 2.24 \\
sf1 & 0.6775 & 0.6905 & \textbf{0.7580} & 1.89 \\
sf2 & 0.4907 & 0.5014 & \textbf{0.6181} & 1.97 \\
slump & 0.4855 & 0.5849 & \textbf{0.6054} & 2.46 \\
wq & 0.6580 & 0.6191 & \textbf{0.7598} & 2.74 \\
\midrule
average & 0.5977 & 0.6326 & \textbf{0.7145} & 2.47 \\
\bottomrule
\end{tabular}
\end{table}

\begin{figure*}[ht]
    \centering
    \begin{subfigure}[t]{0.55\textwidth}
        \centering
        \includegraphics[width=\textwidth]{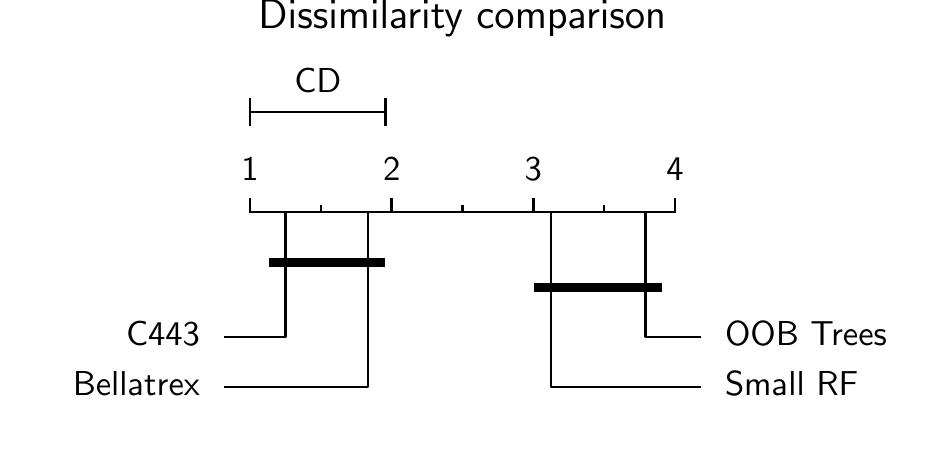}
        {{\small binary data}}    
        \label{fig:dissim-nemenyi-bin}
    \end{subfigure}
    \vfill
    \vspace{0.2cm}
    \begin{subfigure}[b]{0.47\textwidth}
        \centering
        \includegraphics[width=\textwidth]{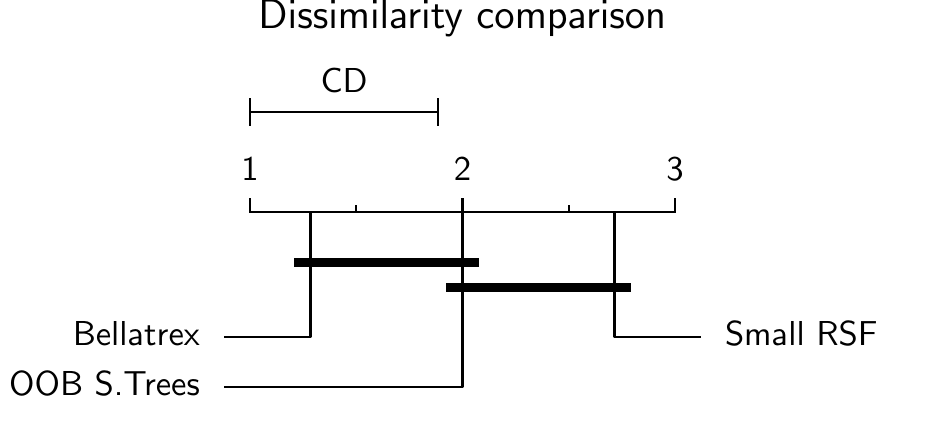}
        \captionsetup{width=.9\linewidth}
        {{\small time-to-event data}}    
        \label{fig:dissim-nemenyi-surv}
    \end{subfigure}
    \hfill
    \begin{subfigure}[b]{0.47\textwidth}  
        \centering 
        \includegraphics[width=\textwidth]{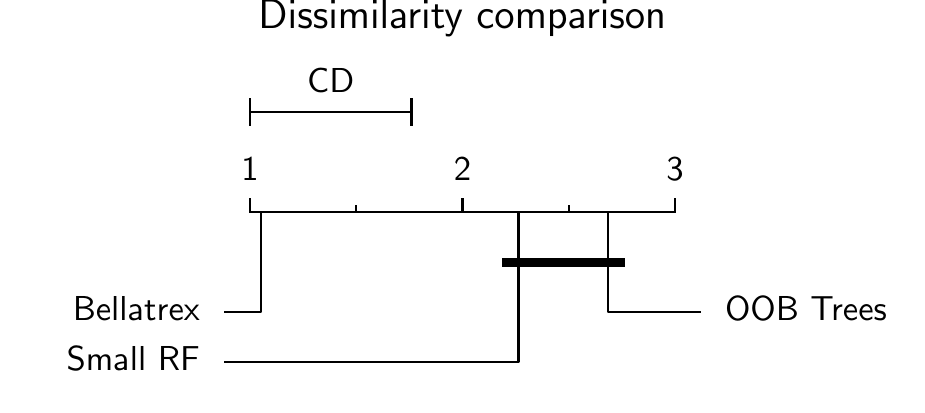}
        {{\small regression tasks}}    
        \label{fig:dissim-nemenyi-regress}
    \end{subfigure}
    \vskip\baselineskip
    \vspace{0.4cm}
    \begin{subfigure}[b]{0.47\textwidth}   
        \centering 
        \includegraphics[width=\textwidth]{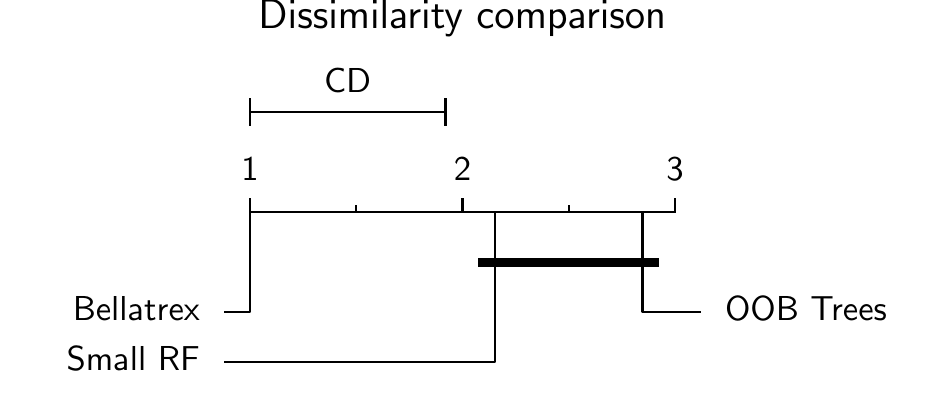}
        {{\small multi-label classification data}}    
        \label{fig:dissim-nemenyi-multi}
    \end{subfigure}
    \hfill
    \begin{subfigure}[b]{0.47\textwidth}   
        \centering
        \includegraphics[width=\textwidth]{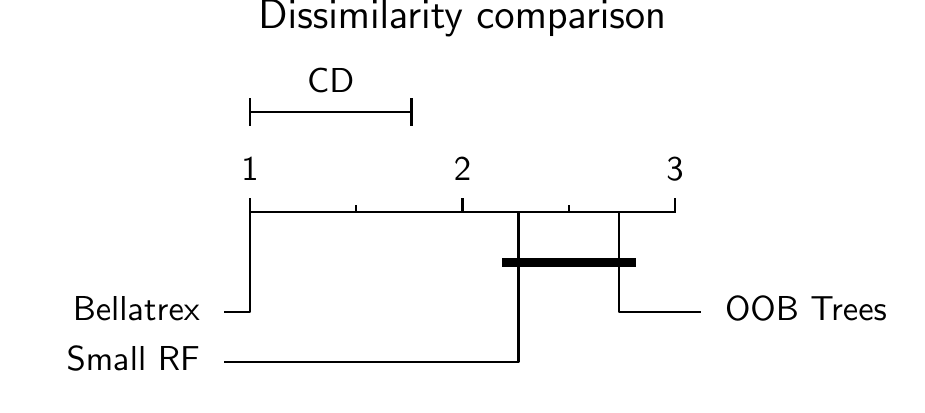}
        {{\small multi-target regression data}}    
        \label{fig:dissim-nemenyi-mtr}
    \end{subfigure}
    \vspace{0.4cm}
    \caption[fig:dissim-nemenyi-all]
    {Nemenyi test statistical test, performed on the average achieved dissimilarities across the five scenarios.} 
    \label{fig:dissim-nemenyi-all}
\end{figure*}

\FloatBarrier
\newpage


\subsection{Ablation studies}

Here, we report the dataset-specific results results for the ablation study, as well as the outcomes of the statistical testing procedure, for time-to-event, multi-label classification, single target regression, and multi-target regression data.  Detailed results on binary classification datasets are reported in the main body, Section~\ref{sec:ablations}.

\begin{table}[ht]
\tiny
\centering
\caption{Ablation study results on the time-to-event datasets. In bold, the best achieved performance.
}
\label{tab:ablation-survival}
\begin{tabular}{lrrrr}
\toprule
 & \makecell{Bellatrex \\ original} & \makecell{no pre\\ selection} & \makecell{no dim. \\ reduction} & \makecell{none of \\the two} \\
\midrule
addicts & \textbf{0.6512} & 0.6411 & 0.6503 & 0.6451 \\
B.C. survival & 0.6417 & 0.6060 & \textbf{0.6440} & 0.6245 \\
DBCD & \textbf{0.7534} & 0.7330 & 0.7072 & 0.5923 \\
DLBCL & \textbf{0.6357} & 0.5762 & 0.6118 & 0.5423 \\
echocardiogram & 0.4166 & 0.4199 & 0.4212 & \textbf{0.4799} \\
FLChain & 0.8330 & \textbf{0.8331} & 0.8317 & 0.8188 \\
gbsg2 & \textbf{0.7035} & 0.7030 & 0.7025 & 0.6776 \\
lung & 0.6236 & \textbf{0.6265} & 0.6175 & 0.5949 \\
NHANES I & 0.8209 & 0.8136 & \textbf{0.8215} & 0.8070 \\
PBC & \textbf{0.8467} & 0.8417 & 0.8457 & 0.8076 \\
rotterdam (excl. rec.) & \textbf{0.7961} & \textbf{0.7961} & 0.7958 & 0.7870 \\
rotterdam (incl. rec.) & 0.9047 & \textbf{0.9069} & 0.9050 & 0.8945 \\
veteran & 0.7349 & 0.7277 & \textbf{0.7389} & 0.7130 \\
whas500 & 0.7458 & 0.7412 & \textbf{0.7466} & 0.7373 \\
\midrule
average & \textbf{0.7224} & 0.7122 & 0.7175 & 0.6946 \\
\bottomrule
\end{tabular}
\end{table}

\begin{table}[ht]
\tiny
\centering
\begin{tabular}{lrrrr}
\toprule
 & \makecell{\textbf{Bellatrex} \\ (original)} & \makecell{no pre\\ selection} & \makecell{no dim. \\ reduction} & \makecell{none of \\the two} \\
\midrule
birds & \textbf{0.8190} & 0.7914 & 0.8184 & 0.7859 \\
CAL500 & \textbf{0.5428} & 0.5375 & 0.5394 & 0.5320 \\
emotions & \textbf{0.8205} & 0.8000 & \textbf{0.8205} & 0.7523 \\
enron & \textbf{0.7707} & 0.7443 & 0.7576 & 0.7154 \\
flags & 0.7237 & 0.6877 & \textbf{0.7317} & 0.6854 \\
genbase & 0.9977 & \textbf{0.9992} & 0.9976 & 0.9984 \\
langlog & \textbf{0.6075} & 0.5848 & 0.5995 & 0.6071 \\
medical & \textbf{0.9497} & 0.9299 & 0.9365 & 0.9137 \\
ng20 & \textbf{0.9225} & 0.9089 & 0.9082 & 0.8491 \\
scene & \textbf{0.9195} & 0.8993 & 0.9171 & 0.8377 \\
slashdot & \textbf{0.8186} & 0.7985 & 0.8035 & 0.7770 \\
stackex chess & \textbf{0.6928} & 0.6819 & 0.6776 & 0.6596 \\
yeast & \textbf{0.6777} & 0.6771 & 0.6623 & 0.6243 \\
\midrule
average & \textbf{0.7894} & 0.7723 & 0.7823 & 0.7491 \\
\bottomrule
\end{tabular}

\caption{Ablation study results on the multi-label classification datasets. In bold, the best achieved performance.}
\label{tab:ablation-multilabel}
\end{table}

\begin{table}[!ht]
\centering
\tiny

\begin{tabular}{lrrrr}
\toprule
 & \makecell{\textbf{Bellatrex} \\ (original)} & \makecell{no pre\\ selection} & \makecell{no dim. \\ reduction} & \makecell{none of \\the two} \\
\midrule
airfoil & \textbf{0.0374} & 0.0389 & 0.0377 & 0.0427 \\
AmesHousing & 0.0230 & 0.0243 & \textbf{0.0228} & 0.0277 \\
auto mpg & \textbf{0.0525} & 0.0545 & 0.0526 & 0.0571 \\
bike sharing & \textbf{0.0481} & 0.0487 & \textbf{0.0481} & 0.0521 \\
boston housing & 0.0576 & 0.0587 & \textbf{0.0569} & 0.0620 \\
california & 0.0611 & 0.0629 & \textbf{0.0610} & 0.0654 \\
car imports & 0.0360 & 0.0367 & \textbf{0.0359} & 0.0403 \\
Computer & 0.0297 & 0.0294 & \textbf{0.0293} & 0.0305 \\
concrete compress & 0.0396 & 0.0418 & \textbf{0.0393} & 0.0449 \\
concrete slump & 0.0734 & 0.0744 & \textbf{0.0727} & 0.0742 \\
ENB2012 cooling & 0.0338 & \textbf{0.0335} & 0.0336 & 0.0338 \\
ENB2012 heating & 0.0111 & 0.0112 & \textbf{0.0110} & 0.0117 \\
forest fires & \textbf{0.3032} & 0.3084 & 0.3044 & 0.3081 \\
PRSA data & 0.0361 & 0.0363 & \textbf{0.0360} & 0.0388 \\
slump dataset & 0.0680 & 0.0688 & \textbf{0.0663} & 0.0730 \\
students maths & 0.1550 & 0.1554 & \textbf{0.1538} & 0.1623 \\
wine quality all & 0.0710 & 0.0711 & \textbf{0.0692} & 0.0748 \\
wine quality red & 0.0808 & 0.0825 & \textbf{0.0776} & 0.0792 \\
wine quality white & 0.0623 & 0.0614 & \textbf{0.0611} & 0.0645 \\
\midrule 
average & 0.0674 & 0.0684 & \textbf{0.0668} & 0.0707 \\
\bottomrule
\end{tabular}
\captionsetup{width=.9\linewidth}
\caption{Ablation study results on the regression datasets. In bold, the best achieved performance.}
\label{tab:ablation-regression}
\end{table}

\begin{table}[!ht]
\centering
\tiny

\begin{tabular}{lrrrr}
\toprule
 & \makecell{\textbf{Bellatrex} \\ (original)} & \makecell{no pre\\ selection} & \makecell{no dim. \\ reduction} & \makecell{none of \\the two} \\
\midrule
andro & 0.0911 & 0.0968 & 0.0899 & \textbf{0.0865} \\
atp1d & 0.0479 & 0.0483 & \textbf{0.0467} & 0.0545 \\
atp7d & 0.0443 & 0.0437 & \textbf{0.0413} & 0.0720 \\
edm & 0.1068 & 0.1083 & \textbf{0.0982} & 0.1037 \\
enb & 0.0225 & 0.0231 & \textbf{0.0224} & 0.0232 \\
ENB2012 & 0.0234 & 0.0236 & \textbf{0.0231} & 0.0240 \\
jura & \textbf{0.0678} & 0.0703 & 0.0687 & 0.0732 \\
oes10 & \textbf{0.0220} & 0.0237 & 0.0225 & 0.0269 \\
oes97 & \textbf{0.0296} & 0.0310 & 0.0298 & 0.0350 \\
osales & \textbf{0.0361} & 0.0376 & 0.0362 & 0.0413 \\
rf1 & \textbf{0.0019} & 0.0022 & \textbf{0.0019} & 0.0025 \\
rf2 & 0.0031 & 0.0033 & \textbf{0.0030} & 0.0043 \\
scm1d & 0.0249 & 0.0264 & \textbf{0.0247} & 0.0288 \\
scm20d & 0.0296 & 0.0307 & \textbf{0.0292} & 0.0353 \\
scpf & 0.0105 & 0.0104 & 0.0105 & \textbf{0.0103} \\
sf1 & \textbf{0.0726} & 0.0731 & 0.0727 & 0.0741 \\
sf2 & 0.0318 & \textbf{0.0313} & \textbf{0.0313} & 0.0316 \\
slump & 0.1433 & 0.1387 & 0.1426 & \textbf{0.1357} \\
wq & 0.1610 & 0.1639 & \textbf{0.1608} & 0.1696 \\
\midrule
average & 0.0511 & 0.0519 & \textbf{0.0503} & 0.0543 \\
\bottomrule
\end{tabular}
\captionsetup{width=.9\linewidth}
\caption{Ablation study results on the multi-target regression datasets. In bold, the best achieved performance.}
\label{tab:ablation-multitarget}
\end{table}

\FloatBarrier

\section*{Acknowledgements}

We thank Niels Liebeer for his help in implementing the UI of the Bellatrex method.

\printbibliography


\end{document}